\documentclass[]{informs3noheader}
\OneAndAHalfSpacedXI

\usepackage[utf8]{inputenc}
\usepackage[english]{babel}

\usepackage{graphicx}     
\usepackage{xcolor}       
\usepackage{comment}
\usepackage{verbatim}
\usepackage{float}
\usepackage{adjustbox}
\usepackage[normalem]{ulem}
\usepackage{soul}
\usepackage{blindtext}
\usepackage{longtable}
\usepackage{lscape}
\usepackage{subcaption}
\usepackage{amssymb}
\usepackage{mathtools}
\usepackage{cases}
\usepackage[mathscr]{euscript}
\usepackage{mathrsfs}
\usepackage{bbm}

\usepackage{array}
\usepackage{tabularx}
\usepackage{multirow}
\usepackage{makecell}
\usepackage{booktabs}
\usepackage[flushleft]{threeparttable}
\usepackage{arydshln}
\usepackage{caption} 
\renewcommand{\arraystretch}{1.0}

\newcolumntype{M}[1]{>{\centering\arraybackslash}p{#1}}

\usepackage[inline]{enumitem}
\setlist[itemize]{left=1.2em, itemsep=0pt, topsep=2pt, parsep=0pt, partopsep=0pt}

\usepackage{algorithm}
\usepackage{algorithmicx}
\usepackage{algpseudocode}

\usepackage{listings}

\lstdefinelanguage{Julia}{
  morekeywords={abstract,break,case,catch,const,continue,do,else,elseif,end,export,false,
      for,function,if,import,in,let,macro,module,mutable,primitive,quote,return,struct,
      true,try,type,using,while},
  sensitive=true,
  morecomment=[l]\#,
  morestring=[b]",
}

\usepackage{tikz}
\usetikzlibrary{shapes.geometric}
\usetikzlibrary{positioning,fit,arrows.meta,backgrounds}
\usetikzlibrary{decorations.pathreplacing}
\usetikzlibrary{calc}

\definecolor{mygreen}      {RGB}{0,90,0}
\definecolor{myblue}       {RGB}{0,51,140}
\definecolor{myorange}     {RGB}{238,118,0}
\definecolor{myred}        {RGB}{126,0,0}
\definecolor{mygray}       {RGB}{100,100,105}
\definecolor{mygrayblue}   {RGB}{0,128,128}
\definecolor{mygraygreen}  {RGB}{128,128,0}
\definecolor{DarkPurple}   {RGB}{142, 36, 170}
\definecolor{LightPurple}  {RGB}{57, 130, 7}

\pgfdeclarelayer{agentbg}
\pgfdeclarelayer{toolkitbg}
\pgfsetlayers{agentbg,toolkitbg,main}

\tikzstyle{user}   = [rectangle, thick, rounded corners, minimum width=3cm, minimum height=1.5cm,
                      text centered, font=\Large, draw=black, fill=green!15]
\tikzstyle{solver} = [rectangle, thick, rounded corners, minimum width=3cm, minimum height=1.5cm,
                      text centered, font=\Large, draw=black, fill=purple!15]
\tikzstyle{llm}    = [rectangle, rounded corners, minimum width=3cm, minimum height=1.5cm,
                      text centered, font=\large, draw=black, fill=yellow!15]
\tikzstyle{prompt} = [rectangle, rounded corners, minimum width=3cm, minimum height=1.5cm,
                      text centered, font=\large, draw=black, fill=blue!15]
\tikzstyle{tools}  = [rectangle, rounded corners, minimum width=3.5cm, minimum height=0.5cm,
                      text centered, font=\small, draw=black, fill=blue!5]
\tikzstyle{arrow}  = [thick,->,>=stealth]
\tikzstyle{block}  = [rectangle, minimum width=4cm, minimum height=1.2cm,
                      text centered, font=\small, draw=black, fill=gray!15]
\tikzstyle{block2}  = [rectangle, minimum width=2cm, minimum height=1.2cm,
                      text centered, font=\small, draw=black, fill=gray!15]

\usepackage{natbib}
\bibpunct[, ]{(}{)}{,}{a}{}{,}%
\def\bibfont{\small}%
\usepackage[most]{tcolorbox}

\tcbuselibrary{breakable,skins}

\definecolor{BubbleUser}{RGB}{240,244,255}   
\definecolor{BubbleAI}{RGB}{244,244,244}     
\definecolor{BubbleFrame}{RGB}{220,220,220}

\newlength{\bubblewidth}
\tcbset{
  chatbubble/.style={
    enhanced, breakable,
    boxrule=0pt,
    colframe=BubbleFrame,
    arc=12pt,
    left=10pt, right=10pt, top=8pt, bottom=8pt,
    before skip=6pt, after skip=6pt,
    fontupper=\fontsize{8}{9.5}\selectfont,
    before upper={\setlength{\parskip}{1pt}\setlength{\parindent}{0pt}},
  }
}

\newtcolorbox{userbubble}[1][]{
  chatbubble,
  colback=BubbleUser,
  overlay={%
    \begin{scope}
      \fill[BubbleUser]
        ([xshift=4pt]frame.south east)
        .. controls +(6pt,4pt) and +(2pt,-2pt)
        .. ([xshift=8pt,yshift=8pt]frame.south east)
        -- cycle;
    \end{scope}
  },
  #1
}

\newtcolorbox{aibubble}[1][]{
  chatbubble,
  colback=BubbleAI,
  overlay={%
    \begin{scope}
      \fill[BubbleAI]
        ([xshift=-4pt]frame.south west)
        .. controls +(-6pt,4pt) and +(-2pt,-2pt)
        .. ([xshift=-8pt,yshift=8pt]frame.south west)
        -- cycle;
    \end{scope}
  },
  #1
}

\newtcolorbox{dialoguebox}[1][]{
  enhanced,
  colback=white,
  colframe=black!20,
  boxrule=0.6pt,
  arc=10pt,
  left=6pt, right=6pt, top=6pt, bottom=6pt,
  before skip=10pt, after skip=10pt,
  #1
}

\tcbset{colback=gray!5, colframe=gray!40, sharp corners, boxsep=1ex}
\newtcolorbox{msg}[2][]{title=\textbf{#2}, #1}

\usepackage{fontawesome5}
\DeclareRobustCommand{\usericon}{\faIcon{user}}
\DeclareRobustCommand{\robicon}{\raisebox{-0.2ex}{\smash{\faIcon{robot}}}}

\usepackage{hyperref}
\hypersetup{
  colorlinks=true,
  linkcolor=blue,
  citecolor=blue,
  filecolor=black,
  urlcolor=black,
}

\usepackage{lineno}

\TheoremsNumberedThrough
\EquationsNumberedThrough
\MANUSCRIPTNO{}

\def\calD{\mathcal{D}}

\def\calI{\mathcal{I}}

\def\calQ{\mathcal{Q}}

\def\calS{\mathcal{S}}
\def\calT{\mathcal{T}}

\def\bc{\boldsymbol{c}}

\def\bi{\boldsymbol{i}}

\def\bx{\boldsymbol{x}}

\def\bgamma{\boldsymbol{\gamma}}

\definecolor{mygreen}    {RGB}{0,90,0}
\definecolor{myblue}     {RGB}{0,51,140}
\definecolor{myorange}   {RGB}{238,118,0}
\definecolor{myred}      {RGB}{126,0,0}
\definecolor{mygray}     {RGB}{100,100,105}
\definecolor{mygrayblue} {RGB}{0,128,128}
\definecolor{mygraygreen}{RGB}{128,128,0}
\definecolor{DarkPurple}     {RGB}{142, 36, 170}
\definecolor{LightPurple}    {RGB}{57, 130, 7}

\begin{document}

\RUNAUTHOR{}

\RUNTITLE{LLM Agents for Interactive Optimization}

\TITLE{Let's Have a Conversation: Designing and Evaluating LLM Agents for Interactive Optimization}

\ARTICLEAUTHORS{
	\AUTHOR{Joshua Drossman, Alexandre Jacquillat}
    \AFF{
        Operations Research Center and Sloan School of Management, Massachusetts Institute of Technology,  Cambridge, MA.
    }
	\AUTHOR{Sébastien Martin}
	\AFF{Kellogg School of Management, Northwestern University, Evanston, IL
	}
}

\ABSTRACT{
    \textbf{Problem definition.} Optimization is as much about modeling the right problem as solving it. Identifying the right objectives, constraints, and trade-offs demands extensive interaction between researchers and stakeholders. Large language models can empower decision-makers with optimization capabilities through interactive optimization agents that can propose, interpret and refine solutions. However, it is fundamentally harder to evaluate a conversation-based interaction than traditional one-shot approaches.
    \textbf{Methodology/results.} This paper proposes a scalable and replicable methodology for evaluating optimization agents through conversations. We build LLM-powered decision agents that role-play diverse stakeholders, each governed by an internal utility function but communicating like a real decision-maker. We generate thousands of conversations in a school scheduling case study. Results show that one-shot evaluation is severely limiting: the same optimization agent converges to much higher-quality solutions through conversations. Then, this paper uses this methodology to demonstrate that tailored optimization agents, endowed with domain-specific prompts and structured tools, can lead to significant improvements in solution quality in fewer interactions, as compared to general-purpose chatbots.
    \textbf{Managerial implications.} These findings provide evidence of the benefits of emerging solutions at the AI-optimization interface to expand the reach of optimization technologies in practice. They also uncover the impact of operations research expertise to facilitate interactive deployments through the design of effective and reliable optimization agents.
}


\maketitle
\vspace{-18pt}

\section{Introduction and Background}

Optimization methods have enabled significant advances across domains such as energy, healthcare, transportation, supply chains, and the public sector. Thanks to substantial progress in computing and algorithms, optimization methods can power significant improvements in operating efficiency, strategic planning, and sustainability. But in practical deployments, the hardest part is often less about solving a model than modeling the right problem---identifying the right objectives, the right constraints, and the right trade-offs to reveal decision-makers’ most preferred solution.

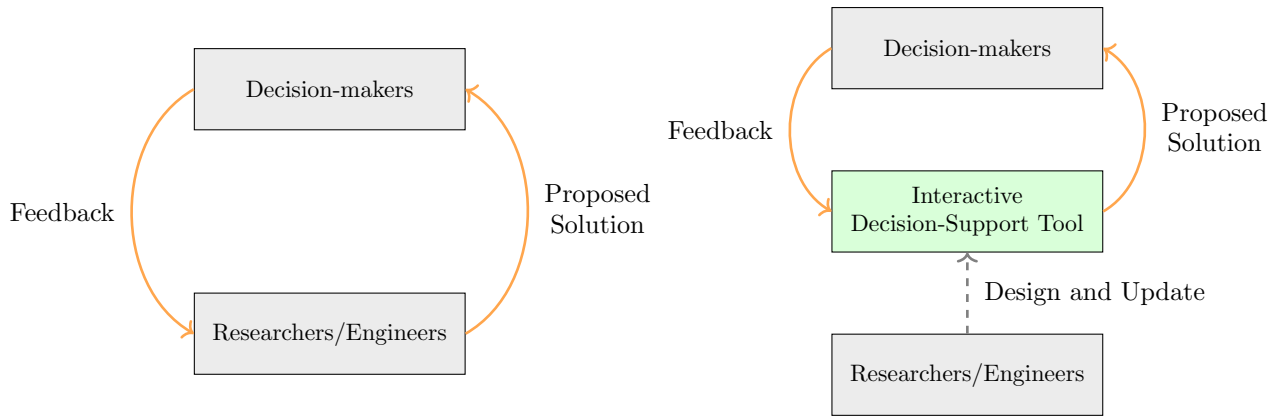
\begin{figure}[h!]
    \centering
    \begin{subfigure}[c]{0.47\textwidth}
        \centering
        \begin{tikzpicture}[
        squarednode/.style={rectangle, draw=black, fill=white, thick, minimum size=15mm},
        scale=0.9,transform shape
        ]
        
        \node (decision_maker) [block] {Decision-makers};
        \node (researcher) [block, below=2.4cm of decision_maker] {Researchers/Engineers};
        \node[minimum height=0.6cm, above=0cm of decision_maker] (spacer) {};
        \node[minimum height=0.6cm, below=0cm of researcher] (spacer) {};
        
        \draw[->, line width=1pt, color=orange!70] 
            ($(decision_maker.west)$) to[out=210,in=150] node[midway, left=3pt, color=black]{Feedback}($(researcher.west)+(0,0)$);
        
        \draw[<-, line width=1pt, color=orange!70] 
            ($(decision_maker.east)$) to[out=-30,in=30] node[midway, right=3pt, color=black]{\makecell{Proposed\\Solution}}($(researcher.east)+(0,0)$);
        \end{tikzpicture}
        \caption{Traditional optimization process.}
        \label{fig:traditional}
    \end{subfigure}
    \hfill 
    \begin{subfigure}[c]{0.47\textwidth}
        \centering
        \begin{tikzpicture}[
        squarednode/.style={rectangle, draw=black, fill=white, thick, minimum size=15mm},
        scale=0.9,transform shape
        ]
        
        \node (decision_maker) [block] {Decision-makers};
        \node (tool) [block, fill=green!15, align=center, below=1.2cm of decision_maker] {\makecell{Interactive\\Decision-Support Tool}};
        \node (researcher) [block, below=1.2cm of tool] {Researchers/Engineers};
        
        \draw[->, line width=1pt, color=orange!70] 
            ($(decision_maker.west)$) to[out=210,in=150] node[midway, left=3pt, color=black]{Feedback}($(tool.west)+(0,0)$);
        
        \draw[<-, line width=1pt, color=orange!70] 
            ($(decision_maker.east)$) to[out=-30,in=30] node[midway, right=3pt, color=black]{\makecell{Proposed\\Solution}}($(tool.east)+(0,0)$);
        
        \draw[->, dashed, line width=1pt, color=gray!100] 
            ($(researcher.north)$) -- node[midway, right=3pt, color=black] {Design and Update} ($(tool.south)+(0,0)$);
        \end{tikzpicture}
        \caption{Interactive optimization with decision tool.}
        \label{fig:interactive}
    \end{subfigure}
    \caption{Visualization of approaches to optimization in practice.}
    \vspace{-12pt}
\end{figure}

This process has traditionally required extensive back-and-forth between decision-makers and operations research experts to develop the ``right'' optimization model, through meetings, prototypes, and iterative refinements (Figure~\ref{fig:traditional}). Even in largely automated systems (e.g., \cite{azagirre_2024_better} in ride-sharing, \cite{allgor_2023_algorithm} in warehousing operations), various teams must first figure out what to optimize. In other sectors where humans retain decision authority, the penetration of optimization solutions remain even more limited---a typical case of ``algorithm aversion'' \cite{dietvorst_overcoming_2018}. Successful deployments are still possible (e.g., \cite{delarue_algorithmic_2024} in public schools, \cite{holland_2017_ups} in logistics, \cite{baxi_2025_online} in data centers, etc.) but require back-and-forth between decision-makers and technical teams to formulate an optimization model, build trust in the algorithm, and iteratively refine the solution based on emerging requirements. These interactions are costly, slow, and limited to organizations with the resources to sustain them.

Natural human-computer interactions enabled by large language models (LLMs) offer a new path forward. LLM-based conversational chatbots allow a decision-maker to interact with an optimization model by playing the role of a researcher in traditional deployments \citep{simchi2025democratizing}. This relies on an LLM-powered \emph{optimization agent} to interact with a mathematical model, build and run code, propose solutions, interpret trade-offs, and refine formulations through dialogue. To be effective, such an interactive optimization process requires more than the technical ability to formulate and solve models correctly. The agent must also understand what the decision-maker wants through conversation. Human decision-makers---especially those without a background in optimization---do not think in terms of objective functions and constraints. Rather, they may describe goals qualitatively, express soft preferences, and even discover new requirements through experimentation and upon seeing concrete proposals. In turn, optimization agents need both the technical ability to formulate and solve a model as well as the practical ability to interact with a decision-maker to elicit preferences iteratively, and so does any framework that evaluates them.

This paper develops optimization agents for interactive decision-making along with a rigorous LLM-enabled assessment methodology. We ground our work in a school scheduling problem from the San Francisco Unified School District (SFUSD) \citep{delarue_algorithmic_2024,jeloudar2025optimization}, where stakeholders (principals, parents, administrators, transportation coordinators) redesign school start times. This setting exemplifies the broader challenge: preferences are diverse, hard to articulate formally, and often discovered through an interactive process. This running example makes every concept concrete throughout the paper; yet, all the developments are generalizable by design. We address two complementary questions. First, how can optimization agents be rigorously evaluated in settings where performance depends on multi-turn interactions? Second, how should optimization agents be designed to help decision-makers reach good solutions over the course of a conversation?

\subsection{LLMs and Optimization}

The most common approach to integrating LLMs with optimization is \textit{autoformulation}, which tasks an LLM with translating a text-based problem description into a mathematical model. The NL4Opt Competition investigated how natural language processing can produce optimization formulations from text \citep{ramamonjison_augmenting_2022,ramamonjison_nl4opt_2023}. \citet{ahmaditeshnizi_optimus_2023} and \citet{xiao_coe_2024} developed multi-agent architectures for formulating and solving mixed-integer programs from natural language, with extensions to large-scale settings \citep{liang_largescale_2026}. \citet{astorga_autoformulation_2024} proposed an LLM-enhanced Monte-Carlo tree search for autoformulation, \citet{huang_orlm_2025} demonstrated the potential of fine-tuning and model alignment, with further improvements by \citet{jiang_llmopt_2025}. Other approaches include chain-of-thought reasoning \citep{yang_expert_2025} and experience learning \citep{kong_alphaopt_2026}. There have also been significant industry developments, with Gurobi's \textit{Gurobot} \citep{quintero_2025_gurobot} and Microsoft Research's \textit{OptiMind} \citep{consulting_optimind_2026}.

\begin{figure}[h!]
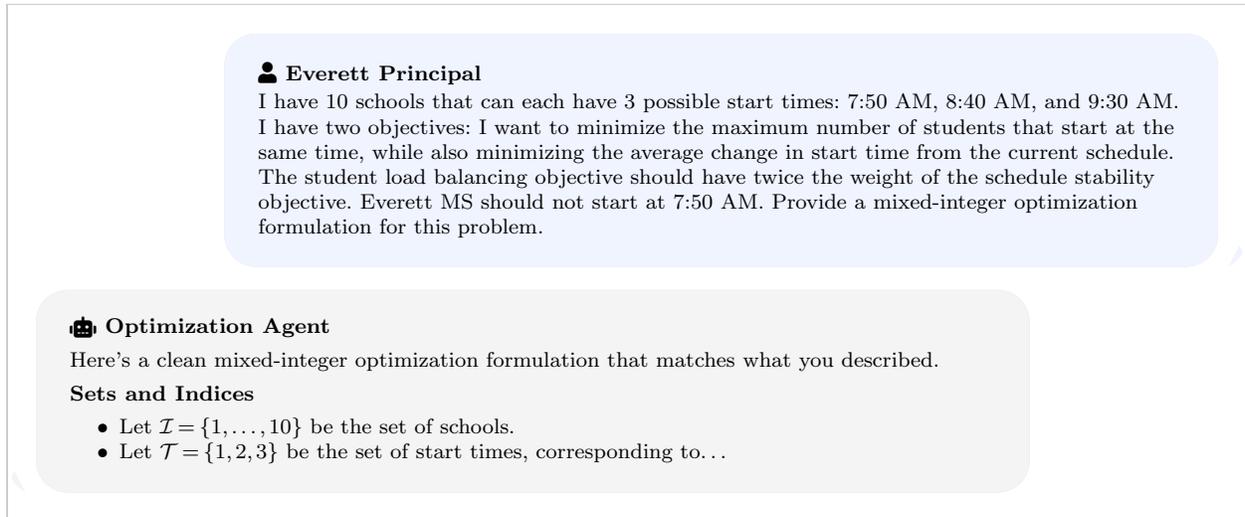

\centering
\begin{dialoguebox}
    \begin{minipage}{\linewidth}
      \noindent\hfill
      \begin{minipage}[t]{\bubblewidth}
        \begin{userbubble}
            \textbf{\usericon\ Everett Principal}\par
            \raggedright
          I have 10 schools that can each have 3 possible start times: 7:50 AM, 8:40 AM, and 9:30 AM. I have two objectives: I want to minimize the maximum number of students that start at the same time, while also minimizing the average change in start time from the current schedule. The student load balancing objective should have twice the weight of the schedule stability objective. Everett MS should not start at 7:50 AM. Provide a mixed-integer optimization formulation for this problem.
        \end{userbubble}
      \end{minipage}

      \vspace{0.8em}
    
      \noindent
      \begin{minipage}[t]{\bubblewidth}
        \begin{aibubble}
        \textbf{\robicon\ Optimization Agent}\par\vspace{2pt}
          \raggedright
            Here's a clean mixed-integer optimization formulation that matches what you described.\\
            \vspace{0.4em}
            \textbf{Sets and Indices}
            \begin{itemize}
                \item Let $\calI=\{1,\dots,10\}$ be the set of schools.
                \item Let $\calT=\{1,2,3\}$ be the set of start times, corresponding to\dots
            \end{itemize}
        \end{aibubble}
      \end{minipage}
    \end{minipage}
\end{dialoguebox}
\caption{Example of LLM-based autoformulation: a school principal describes an optimization problem in precise technical language and the tool returns a mathematical formulation of the problem.}
\label{fig:autoformulation-dialogue}
\vspace{-8pt}
\end{figure}

\begin{figure}[!h]
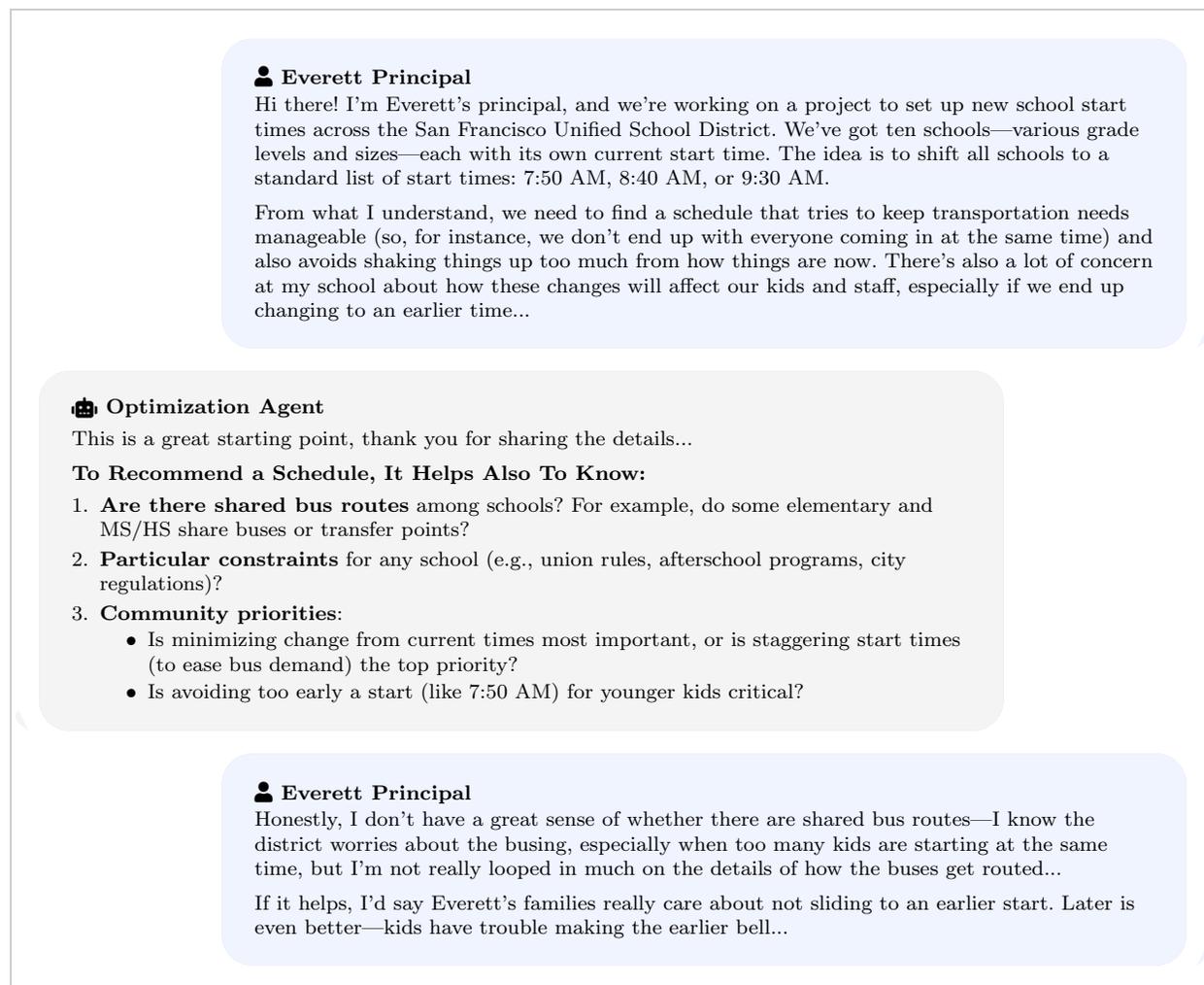

\centering
\begin{dialoguebox}
    \begin{minipage}{\linewidth}
        \noindent\hfill
        \begin{minipage}[t]{\bubblewidth}
            \begin{userbubble}
                \textbf{\protect\usericon\ Everett Principal}\par
                \raggedright
                Hi there! I'm Everett's principal, and we're working on a project to set up new school start times across the San Francisco Unified School District. We've got ten schools—various grade levels and sizes—each with its own current start time. The idea is to shift all schools to a standard list of start times: 7:50 AM, 8:40 AM, or 9:30 AM.\\
                \vspace{4pt}
                From what I understand, we need to find a schedule that tries to keep transportation needs manageable (so, for instance, we don't end up with everyone coming in at the same time) and also avoids shaking things up too much from how things are now. There's also a lot of concern at my school about how these changes will affect our kids and staff, especially if we end up changing to an earlier time...
            \end{userbubble}
        \end{minipage}
        
        \vspace{0.8em}
        
        \noindent
        \begin{minipage}[t]{\bubblewidth}
            \begin{aibubble}
                \textbf{\protect\robicon\ Optimization Agent}\par
                \vspace{2pt}
                \raggedright
                This is a great starting point, thank you for sharing the details...\\
                \vspace{4pt}
                \textbf{To Recommend a Schedule, It Helps Also To Know:}
                \vspace{2pt}
                \begin{enumerate}[nosep, leftmargin=*, itemsep=2pt]
                    \item \textbf{Are there shared bus routes} among schools? For example, do some elementary and MS/HS share buses or transfer points?
                    \item \textbf{Particular constraints} for any school (e.g., union rules, afterschool programs, city regulations)?
                    \item \textbf{Community priorities}:
                    \vspace{1pt}
                    \begin{itemize}[nosep, leftmargin=*, itemsep=1pt]
                        \item Is minimizing change from current times most important, or is staggering start times (to ease bus demand) the top priority?
                        \item Is avoiding too early a start (like 7:50 AM) for younger kids critical?
                   \end{itemize}
               \end{enumerate}
            \end{aibubble}
        \end{minipage}
        
        \vspace{0.8em}
        
        \noindent\hfill
        \begin{minipage}[t]{\bubblewidth}
            \begin{userbubble}
                \textbf{\protect\usericon\ Everett Principal}\par
                \raggedright
                Honestly, I don't have a great sense of whether there are shared bus routes—I know the district worries about the busing, especially when too many kids are starting at the same time, but I'm not really looped in much on the details of how the buses get routed...\\
                \vspace{4pt}
                If it helps, I'd say Everett's families really care about not sliding to an earlier start. Later is even better—kids have trouble making the earlier bell...
            \end{userbubble}
        \end{minipage}
    \end{minipage}
\end{dialoguebox}
\vspace{-8pt}
\caption{Example of a conversational interaction: the principal is far less precise in their goals and requirements and the tool navigates the problem's landscape to guide them to a good solution.}
\label{fig:interactive-dialogue}
\vspace{-8pt}
\end{figure}

These approaches can be effective when the decision-maker can describe the problem in precise, near-mathematical terms, reflected in autoformulation benchmarks like NLP4LP \citep{ahmaditeshnizi_optimus_2023}, IndustryOR \citep{huang_orlm_2025}, MAMO \citep{huang_mamo_2025}, and OptiBench \citep{yang_optibench_2025}. Figure~\ref{fig:autoformulation-dialogue} shows an example from our SFUSD setting in which a school principal specifies objectives, weights, and constraints and the optimization agent formulates an appropriate model. In practice, this can be achieved when the decision-maker themselves has technical training or prior exposure to optimization modeling. But other stakeholders typically describe their needs very differently. In Figure~\ref{fig:interactive-dialogue}, the principal describes concerns about disruption and earlier start times affecting their kids and staff---a typical qualitative consideration that needs to be interpreted in quantitative terms to be incorporated into the model. Fundamentally, autoformulation assumes a close mapping between the text description and the intended formulation. When the description is ambiguous, incomplete, or evolving, there is no mechanism to discover what the decision-maker actually wants, and no way to evaluate whether the formulation captures it ``correctly.''

A growing body of work moves beyond one-shot formulation toward richer interaction with optimization models. \citet{chen_diagnosing_2023} developed a chatbot that diagnoses infeasibility. \citet{li_large_2023} built \textit{OptiGuide} to answer what-if queries about optimization outcomes. \citet{zhang_decision_2025} extended this to sensitivity analyses and explanations through a multi-agent architecture. These tools let decision-makers ask questions about an existing model, but do not support the iterative process of building a formulation from informal preferences. Closer to our work, \citet{lawless_i_2024} developed an interactive tool for meeting scheduling that elicits requirements through critique of earlier suggestions, and \citet{ghiani_integrating_2024} proposed a human-in-the-loop process where decision-makers iteratively refine goals and constraints using natural language. \cite{simchi2025democratizing} described how AI-powered conversational chatbots can democratize access to optimization via a ``4I's'' framework: insight (extracting relevant information from data streams), interpretability (explaining the model's logic and solution), interactivity (enabling human-in-the-loop engagement), and improvisation (supporting adjustments in dynamic business environments).

Turning to the evaluation, existing approaches reflect a one-shot paradigm. \citet{li_large_2023} evaluated \textit{OptiGuide} based on outcome correctness. \citet{zhang_modeling_2024} added explanation correctness using an LLM-as-a-judge. \citet{lawless_i_2024} proposed metrics for code compilation and constraint satisfaction. \citet{ghiani_integrating_2024} relied on subjective judgment to assess preference incorporation. In all cases, evaluations either assume a well-specified problem or examine queries in isolation, without measuring whether the agent can guide a decision-maker over a conversation.

This discussion identifies a research gap in evaluating the outcome of a full interaction between a decision-maker and an optimization agent. As mentioned earlier, an optimization agent must manage both technical aspects---modeling, implementing and solving an optimization problem---and practical aspects---explaining trade-offs while eliciting qualitative and ill-formed preferences from decision-makers. Prior work at the AI-optimization interface has focused on the former by evaluating the agent's ability to solve a well-specified problem; to our knowledge, there is no well-established method to evaluate an agent's ability to elicit preferences and adjust a solution iteratively over an interactive dialogue with a decision-maker. Ideally, this evaluation process could be performed via a large-scale study with human decision-makers, who would score the outcome of each interaction. Unfortunately, this evaluation approach would be expensive, time consuming and problem specific. We instead propose LLM-based \emph{decision agents} that role-play stakeholders \citep{jones_large_2025}. Because these agents are fully specified and reproducible, any researcher can perform an apples-to-apples comparison between optimization agents.

\subsection{Contributions}

\begin{figure}[h!]
\centering
\resizebox{\textwidth}{!}{
\begin{tikzpicture}[
squarednode/.style={rectangle, draw=black, fill=white, thick, minimum size=15mm},
]

\node (agent) [block2] {Optimization Agent};
\coordinate (humans_mid) at ([xshift=3cm]agent.east);

\node (human_1) [block2, above=1.4cm of humans_mid] {Query 1};
\node (human_2) [block2, above=0.1cm of humans_mid] {Query 2};
\node (dots) [below=0.2cm of humans_mid] {\vdots};
\node (human_n) [block2, below=1.4cm of humans_mid] {Query $n$};

\draw[->, thick] 
    ($(human_1.west) + (0,0)$) -- ($(agent.east)+(0,0.15)$);
\draw[->, thick]
    ($(human_2.west) + (0,0)$) -- ($(agent.east)+(0,0)$);
\draw[->, thick]
    ($(human_n.west) + (0,0)$) -- ($(agent.east)+(0,-0.1)$);

\coordinate (panel1_mid) at ($(agent.west)!0.5!(human_n.east)$);
\node[above=4cm of panel1_mid, font=\bfseries] (label1) {One-Shot Evaluation};

\draw[dashed] ([yshift=2.2cm]agent.north east) ++(4.5,0) -- ++(0,-5.8cm);

\node (agent_2) [block2, right=2cm of humans_mid] {Optimization Agent};
\coordinate (humans_mid_2) at ([xshift=3cm]agent_2.east);

\node (human_21) [block2, above=1.4cm of humans_mid_2] {Human 1};
\node (human_22) [block2, above=0.1cm of humans_mid_2] {Human 2};
\node (dots) [below=0.2cm of humans_mid_2] {\vdots};
\node (human_2n) [block2, below=1.4cm of humans_mid_2] {Human $n$};

\draw[<->, thick] 
    ($(human_21.west) + (0,0)$) -- ($(agent_2.east)+(0,0.15)$);
\draw[<->, thick]
    ($(human_22.west) + (0,0)$) -- ($(agent_2.east)+(0,0)$);
\draw[<->, thick]
    ($(human_2n.west) + (0,0)$) -- ($(agent_2.east)+(0,-0.1)$);

\coordinate (panel2_mid) at ($(agent_2.west)!0.5!(human_2n.east)$);
\node[above=4cm of panel2_mid, font=\bfseries] (label1) {Empirical Evaluation};

\draw[dashed] ([yshift=2.2cm]agent_2.north east) ++(4.5,0) -- ++(0,-5.8cm);

\node (agent_3) [block2, right=2cm of humans_mid_2] {Optimization Agent};
\coordinate (humans_mid_3) at ([xshift=3cm]agent_3.east);

\node (human_31) [block2, fill=green!15, above=1.4cm of humans_mid_3] {Decision Agent 1};
\node (human_32) [block2, fill=green!15, above=0.1cm of humans_mid_3] {Decision Agent 2};
\node (dots) [below=0.2cm of humans_mid_3] {\vdots};
\node (human_3n) [block2, fill=green!15, below=1.4cm of humans_mid_3] {Decision Agent $n$};

\draw[<->, thick] 
    ($(human_31.west) + (0,0)$) -- ($(agent_3.east)+(0,0.15)$);
\draw[<->, thick]
    ($(human_32.west) + (0,0)$) -- ($(agent_3.east)+(0,0)$);
\draw[<->, thick]
    ($(human_3n.west) + (0,0)$) -- ($(agent_3.east)+(0,-0.1)$);

\coordinate (panel3_mid) at ($(agent_3.west)!0.5!(human_3n.east)$);
\node[above=4cm of panel3_mid, font=\bfseries] (label1) {Our Approach};
\end{tikzpicture}
}
\caption{LLM-optimization evaluation frameworks. One-shot evaluation: an optimization agent creates a formulation based on one question, and aims to match a ground truth. Empirical evaluation: an optimization agent is scored based on interactions with human decision-makers. Our approach: an optimization agent is scored based on interactions with decision agents based on verifiable and rigorous criteria.}
\label{fig:assessment}
\vspace{-12pt}
\end{figure}
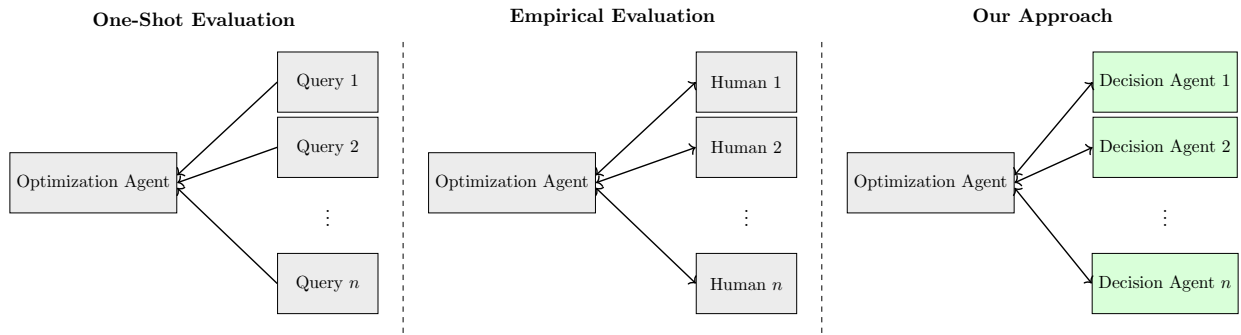

As AI makes optimization increasingly accessible to non-experts, it becomes critical to evaluate whether these tools actually help decision-making. Today, anyone can bring a decision problem to a general-purpose chatbot and get a reasonable attempt at a solution. But real-world optimization is conversational, and existing benchmarks, which test well-specified, one-shot problems, are not properly suited to assess whether an optimization agent performs well over the course of an interaction. Thus, our first contribution is a methodology for evaluating optimization agents through multi-turn conversations. We propose LLM-based decision agents that role-play diverse stakeholders, each governed by an internal utility function but communicating like a human (Figure~\ref{fig:assessment}). This methodology enables controlled, replicable evaluation at scale: in our SFUSD case study, we generate nearly 3,500 conversations between optimization agents and decision agents with varied preferences, communication styles, and levels of self-awareness. Using this framework, we show that one-shot evaluation is severely limiting: the same optimization agent converges to a much higher-quality solution over the course of a conversation than over a one-shot benchmark; in particular, interactive conversations yield a utility-maximizing solution 65\% of the time versus 17\% in a single exchange, even when the decision agent tries to share all relevant information upfront.

Our second contribution is to leverage this evaluation methodology in order to guide the design of optimization agents. A decision-maker could simply use a general-purpose chatbot like ChatGPT or Claude, which is already capable of formulating optimization problems, writing code, and iterating on solutions. Alternatively, an operations researcher could design a specialized agent with domain-specific prompts, structured tools, and solution processing tailored to the problem at hand. Is this additional effort worth it? We answer affirmatively by comparing four agent designs with increasing degrees of structure. Specifically, careful design of an optimization agent addresses distinct failure modes of general-purpose chatbots, such as enumeration without an optimization solver, inefficient interaction with a decision-maker, and silent modeling errors through free-form code. These findings hold across model generations (from GPT-3.5 to GPT-5), though the gap narrows as foundation models improve. Ultimately, these results show the benefits of prompting, solution processing, and dedicated tools in the design of an optimization agent for the SFUSD school scheduling problem.

In summary, this paper develops an LLM-based methodology for assessing conversation-based interactions between a decision-maker and an optimization agent; it provides evidence of the benefits of emerging solutions at the AI-optimization interface to expand the reach of optimization technologies in practice; and it identifies the role of operations research expertise to design effective and reliable optimization agents. In the remainder of this paper, we introduce the SFUSD school scheduling problem (Section~\ref{sec:sfusd_example}), design our optimization agents (Section~\ref{sec:optimization}), develop our evaluation methodology based on decision agents (Section~\ref{sec:assessment}), and present empirical results (Section~\ref{sec:results}).

\section{Running Example: School Scheduling in San Francisco}
\label{sec:sfusd_example}

The framework developed in this paper applies broadly to any setting where decision-makers use optimization to support complex decisions. However, the concepts of optimization agents, decision agents, utility functions, and conversational evaluation can feel abstract without a concrete problem. For this reason, we anchor them in a running example that we use throughout the paper.

We use the problem of redesigning school start times in the SFUSD, studied by  \citep{delarue_algorithmic_2024}. In this setting, the main challenge was not computational but practical. A previous optimization-driven attempt to change school start times in Boston failed because stakeholders felt excluded from the process. In San Francisco, the authors instead built interactive tools that allowed stakeholders (principals, parents, administrators, transportation coordinators) to navigate the solution space, explore the trade-offs between competing objectives, discover what was feasible, and converge on a schedule that addressed their concerns. The lesson is that placing optimization in the hands of people affected by the decisions can greatly facilitate the adoption of optimization.

This setting is well suited for studying interactive optimization. At a high level, it involves two competing objectives: minimizing transportation costs and minimizing disruption to families. At the same time, stakeholders have many other preferences: a parent may care primarily about their child's school start time, a transportation coordinator focuses on keeping costs down, and a principal may worry about conflicts with after-school programs. Moreover, these stakeholders may not think in optimization terms like objective weights or constraint bounds. They think ``my kid's school shouldn't start before 8:30'' and ``try not to change things too much for everyone else.'' These features make preference elicitation just as important as solving the optimization problem itself. \citet{delarue_algorithmic_2024} achieved this through interactive spreadsheet tools designed by the research team. In this paper, we investigate instead the use of LLM-based optimization agents.

We adopt a simplified version of the SFUSD problem. The goal is to assign each of 10 schools (listed in Section~\ref{app:school_data}) to one of three standardized start times: 7:50~AM, 8:40~AM, or 9:30~AM. The main quantitative concerns are \emph{load balancing} (spreading students across time slots to reduce transportation costs) and \emph{schedule stability} (minimizing changes from the current schedule to limit disruptions). The 10 schools comprise 5,857 students from pre-K to high school, currently starting at six different times between 8:00~AM and 9:30~AM. We focus on a smaller instance because our evaluation methodology (Section~\ref{sec:assessment}) requires running thousands of multi-turn conversations between LLM-based agents, each involving multiple calls to an optimization solver. Yet the problem remains rich enough for meaningful interaction: with $3^{10}=59{,}049$ possible start time assignments, the solution space is far too large for a stakeholder to explore without the aid of optimization software. Moreover, this setting constitutes a conservative approach in our inquiry, in that the benefits of structured interactions between a decision agent and an optimization agent are expected to become more significant as the problem becomes larger and more combinatorially complex.

\section{Decision Problem and Optimization Agent}
\label{sec:optimization}

We now formalize the decision problem (Section~\ref{subsec:decisionproblem}) and the design of optimization agents (Section~\ref{subsec:optimization_agent}). We then propose an experimental setup across four optimization agents (Section~\ref{subsec:designing}).

\subsection{Decision Problem}
\label{subsec:decisionproblem}

An optimization agent could be used for many tasks: generating visualizations, exploring a model's sensitivity to parameters, or explaining trade-offs between objectives without ever committing to a course of action. In this paper, we focus on a specific but broadly applicable setting: a decision-maker needs to make a decision, and the optimization agent helps them find a good one. This restriction still encompasses a wide range of practical situations (from choosing a school schedule to configuring a supply chain) while providing the structure needed for rigorous evaluation.

Formally, let $\calS$ denote the set of all feasible decisions available to the decision-maker. In SFUSD, let $\calQ$ be the set of schools and $\calT$ the set of allowed start times. We define binary variables $x_{q,t}$ equal to 1 if school $q \in \calQ$ starts at time $t \in \calT$. The decision space is:
\[
    \calS = \left\{\boldsymbol{x} : x_{q,t} \in \{0,1\},\ \sum_{t \in \calT} x_{q,t} = 1,\ \forall q \in \calQ\right\}
\]
Each element of $\calS$ is a complete assignment of start times across the district. The only hard constraint is that each school gets exactly one start time. Preferences, objectives, and soft constraints are not part of $\calS$; they belong to the decision-maker.

Importantly, $\calS$ represents the space of decisions that the decision-maker cares about, not the decision space or feasible region of any optimization model developed through the process. During a conversation, the agent may introduce auxiliary variables, add constraints to encode the decision-maker's preferences, or reformulate parts of the model to improve solvability. These internal representations can differ from $\calS$, but the decision-maker's choice always lies in $\calS$.

We model the decision-maker's preferences through a utility function $U(s)$ that assigns a value to each feasible decision $s \in \calS$. The decision-maker's implicit goal is to choose $s \in \calS$ that maximizes $U(s)$. In general, $U$ can take any form, and our framework does not require any specific structure. To ground the discussion, we now define a specific class of utility functions in the SFUSD example depending on three features of a schedule: school-specific start times, load balancing across time slots, and schedule stability. The former feature corresponds to ``local'' elements based on individual-level decisions whereas the latter two features correspond to ``global'' elements based on system-wide performance. The approach extends beyond the SFUSD case; the key idea is simply that decision-makers have preferences over solution features that can be combined in different ways.

Let $h_t$ denote the actual start time associated with option $t \in \calT$ (with $h_1 = \text{7:50~AM}$, $h_2 = \text{8:40~AM}$, $h_3 = \text{9:30~AM}$), let $o_q$ denote the current start time of school $q \in \calQ$, and let $e_q$ denote the enrollment at school $q$. For any schedule $s \in \calS$, we can compute the two global objectives:
\begin{itemize}
    \item[--] The absolute change in start time for each school: $c_q(s) = \left|\sum_{t \in \calT} h_t x_{q,t} - o_q\right|$, for each $q \in \calQ$. This measures disruption to families, teachers, and after-school programs when a school's start time changes. The district-wide average change $\frac{1}{|\calQ|}\sum_{q \in \calQ} c_q(s)$ captures schedule stability.
    \item[--] The maximum student load across all time slots: $\ell(s) = \max_{t \in \calT} \sum_{q \in \calQ} e_q x_{q,t}$. This measure serves as a proxy for transportation costs: when many students start at the same time, the district needs a larger bus fleet. Minimizing $\ell(s)$ is the load balancing objective.
\end{itemize}

In addition, decision-makers have preferences about individual schools' start times. A parent might care more about their own child's school starting at a convenient time; a principal might want to avoid a particular time slot. We represent these through preference parameters $\gamma_{q,t}$, where a higher value means the decision-maker prefers school $q$ to start at time $t$.

If a decision-maker could fully articulate their preferences, their ideal schedule would solve:
\begin{align}
    \min_{\bx, \bc, \ell} \quad & \alpha \ell + \beta \frac{1}{|\calQ|}\sum_{q\in\calQ}{c_q} - \sum_{q \in \calQ} \sum_{t \in \calT} \gamma_{q,t}\, x_{q,t} \label{eq:dm_obj} \\
    \text{s.t.} \quad & \sum_{t\in\calT}{x_{q,t}} = 1, \quad && \forall q\in\calQ \label{eq:dm_assign} \\
    & c_q \geq \sum_{t\in\calT}{h_tx_{q,t}}-o_q, \quad && \forall q\in\calQ \label{eq:dm_change1} \\
    & c_q \geq o_q-\sum_{t\in\calT}{h_tx_{q,t}}, \quad && \forall q\in\calQ \label{eq:dm_change2} \\
    & \ell \geq \sum_{q\in\calQ}{e_qx_{q,t}}, \quad && \forall t\in\calT \label{eq:dm_load} \\
    & x_{q,t} \in \{0,1\}, \quad && \forall q\in\calQ,t\in\calT \label{eq:dm_binary}
\end{align}
The parameters $\alpha \geq 0$ and $\beta \geq 0$ control how much the decision-maker cares about load balancing versus schedule stability, and $\gamma_{q,t}$ captures their school-specific time preferences. Equation~\eqref{eq:dm_obj} trades off these three concerns; Constraints~\eqref{eq:dm_assign}--\eqref{eq:dm_binary} define the feasible set and the auxiliary variables that track schedule changes (variable $c_q$) and maximum student load (variable $\ell$).

This formulation defines the space of utility functions. Each decision-maker is characterized by a different set of parameters $(\alpha, \beta, \boldsymbol{\gamma})$. A disruption-averse parent might have high $\beta$ and strong $\gamma$ values for their child's school; a cost-focused transportation coordinator might have high $\alpha$ and care comparatively little about individual school times. The space is rich enough to capture a wide variety of stakeholder profiles, yet structured enough for systematic evaluation.

By design, this formulation captures the decision-maker's underlying thought process, rather than a model that would actually be written down or solved explicitly. The optimization agent does not know the actual values of the weights $(\alpha, \beta, \boldsymbol{\gamma})$. In fact, the decision-maker oftentimes does not express these preferences formally---stakeholders frequently discover what trade-offs they are willing to accept only after seeing a concrete proposal \citep{delarue_algorithmic_2024}. For example, a parent may not realize the load balancing concerns until they see a schedule where many schools start at the same time; a transportation coordinator may not know how much schedule disruption they can tolerate until they see the cost savings; etc. The process of interacting with the optimization agent contributes to preference discovery, which makes the interactive setting challenging and valuable.

\subsection{Designing Optimization Agents}
\label{subsec:optimization_agent}

Recent progress in large language models has produced increasingly capable AI agents that go beyond answering questions to actively using tools and interacting with external software. Coding agents, for instance, can write, execute, and debug programs autonomously. In the same spirit, an optimization agent is an LLM that can formulate an optimization model, edit constraints and objectives, call solvers, and interpret results, all through conversation with a decision-maker (see Figure~\ref{fig:interactive-dialogue}).

We model an interaction between a decision-maker and an optimization agent $\omega \in \Omega$ as a conversation $M(\omega)$, after which the decision-maker selects a decision $s(\omega) \in \calS$. This decision depends on both the decision-maker's own preferences and the recommendations that emerged during the conversation. An optimization agent is successful if it consistently helps decision-makers reach high-quality decisions (i.e., high $U(s(\omega))$) while keeping the conversational effort reasonable.

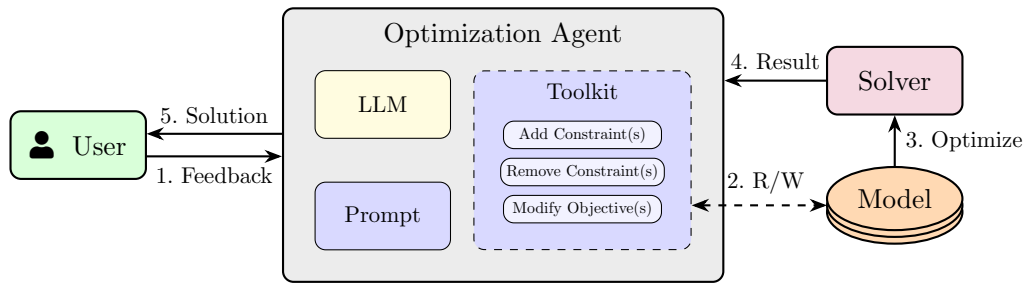
\begin{figure}[h!]
\centering
\begin{tikzpicture}[
  >=Stealth,
  node distance=1.2cm and 1.2cm,
  squarednode/.style={rectangle, draw=black, fill=white, thick, minimum height=10mm, minimum width=20mm},scale=0.6,transform shape
]

\node (tool2) [tools] {Remove Constraint(s)};
\node (tool1) [tools, above=0.2cm of tool2] {Add Constraint(s)};
\node (tool3) [tools, below=0.2cm of tool2] {Modify Objective(s)};
\node (tooltitlepad) [above=0.25cm of tool1] {};

\begin{pgfonlayer}{toolkitbg}
    \node (toolsbox) [
        draw=black, dashed, rounded corners, fill=blue!15, fit={(tooltitlepad) (tool1) (tool2) (tool3)}, inner sep=10pt] {};
\end{pgfonlayer}
\node [font=\large, align=center, anchor=north, yshift=-4pt]
  at (toolsbox.north) {Toolkit};
\node (agenttitlepad) [above=0.4cm of toolsbox] {};

\def\xgap{-0.5cm}

\node (llm) [llm, anchor=north east]
  at ($(toolsbox.north west)+(\xgap,0)$) {LLM};
\node (prompt) [prompt, anchor=south east]
  at ($(toolsbox.south west)+(\xgap,0)$) {Prompt};


\begin{pgfonlayer}{agentbg}
    \node (agentbox) [
      draw=black, thick, rounded corners, fill=gray!15,
      fit={(agenttitlepad) (llm) (prompt) (toolsbox)},
      inner sep=12pt
    ] {};
\end{pgfonlayer}

\node [font=\Large, align=center, anchor=north, yshift=-6pt]
  at (agentbox.north) {Optimization Agent};

\node (user) [user, left=3cm of agentbox] {\faUser\ \ User};

\coordinate (agent-right) at ($(agentbox.east)+(0,0)$);
\coordinate (agent-left-up) at ($(agentbox.west)+(0,0.9)$);
\coordinate (agent-left-down) at ($(agentbox.west)+(-0, -0.9)$);

\coordinate (modelcenter) at ($(toolsbox.east)+(4.5,-1)$);
\begin{scope}[shift={(modelcenter)}]
  \draw[thick, fill=orange!30] (0,-0.15) ellipse (1.5cm and 0.7cm);
  \draw[thick, fill=orange!30] (0, 0.00) ellipse (1.5cm and 0.7cm);
  \draw[thick, fill=orange!30] (0, 0.15) ellipse (1.5cm and 0.7cm);
  \node[text width=3cm, font=\Large, align=center] at (0, 0.15) {Model};
\end{scope}
\coordinate (model) at (modelcenter);

\node (solver) [solver, above=2cm of model] {Solver};


\draw[->, thick]
  ($(user.east)+(0,-0.25)$) -- ($(agentbox.west)+(0,-0.25)$)
  node[midway, below=3pt, font=\large] {1. Feedback};

\draw[->, thick]
  ($(agentbox.west)+(0,0.25)$) -- ($(user.east)+(0,0.25)$)
  node[midway, above=3pt, font=\large] {5. Solution};

\draw[<->, thick, dashed]
  ($(toolsbox.east)+(0,-1)$) -- ($(model)+(-1.5cm,0)$)
  node[midway, above=3pt, xshift=4pt, font=\large] {2. R/W};

\draw[->, thick]
  ($(model)+(0,0.85)$) -- ($(solver.south)+(0,0)$)
  node[midway, right=3pt, font=\large] {3. Optimize};

\draw[->, thick]
  ($(solver.west)+(0,0)$) -- ($(agentbox.east)+(0,1.43)$)
  node[midway, above=3pt, font=\large] {4. Result};
\end{tikzpicture}
\caption{The optimization agent's architecture includes an LLM, prompt, and toolkit, allowing the agent to read/write (R/W) from/to the model, pass the model to the solver, derive its output, and summarize the solution for the decision-maker. This illustration corresponds to the T-P-P design.}
\label{fig:architecture}
\vspace{-12pt}
\end{figure}

There are many ways to build an optimization agent. The design choices range from which LLM to use, to what instructions it receives, to how much freedom it has over the underlying optimization model. These choices, as we will see in Section~\ref{sec:results}, have significant consequences for performance. In this paper, we parameterize an optimization agent by: (i) the foundation model (e.g., GPT-5, Claude Sonnet 4.5, or Gemini 2.5, at the time of our experiments); (ii) the prompt, a set of instructions and background information that shapes the agent's behavior; and (iii) the toolkit, a set of callable functions through which the agent interacts with the optimization model. While one could also specialize an LLM for this task through fine-tuning (e.g., \citet{consulting_optimind_2026}) and Retrieval Augmented Generation, we focus on prompting because it allows for rapid and relatively inexpensive experimentation. Figure~\ref{fig:architecture} illustrates this architecture.

The interplay between prompt and toolkit determines how effectively the agent translates natural-language feedback into formal changes to the optimization model.

\paragraph{Prompt.} The prompt is what transforms a general-purpose LLM into an optimization agent. It is the set of instructions and contextual information that the agent reads before any interaction with the decision-maker. A complete example is provided in Section~\ref{app:opt-assistant-prompt}. We structure our prompt into four components: (i) a task overview providing general instructions about the agent's role, independent of any specific problem; (ii) a problem setting describing the decision problem at hand (in our SFUSD context: the school district, the schools, and the base optimization formulation); (iii) dialogue instructions governing how the agent interacts with decision-makers; and (iv) tool usage instructions specifying when and how to invoke the available tools.

The dialogue instructions deserve particular emphasis. In interactive optimization, the agent is not only a solver but also an interlocutor: it must decide when to ask clarifying questions, when to propose a solution, when to challenge assumptions, and how to present trade-offs to a decision-maker. Should the agent show one solution at a time or compare alternatives? Should it probe for more information before solving, or present a first result and iterate from there? The dialogue instructions encode this interaction style, which can matter as much as the underlying formulation.

Separating general from problem-specific instructions makes it easy to adapt the agent to a new domain by replacing the problem setting alone. Similarly, isolating the dialogue instructions allows us to test how different interaction styles affect performance without changing anything else.

\paragraph{Toolkit.} The toolkit defines what the optimization agent can actually do. While the prompt governs reasoning and communication, the tools control how the agent interacts with the optimization model, and in particular how much freedom it has over the formulation and execution environment.

At one extreme, an agent can write arbitrary code to build and modify the model directly. This provides maximum flexibility, but also risk: the agent may introduce syntax errors, runtime exceptions, or bugs that are hard to detect. At the other extreme, an agent interacts with the model only through a restricted set of predefined functions, each performing a specific, verified operation. This limits what the agent can do but ensures that every modification is valid by construction. The main challenge in designing optimization agents is to define tools that are structured enough to ensure reliable optimization, yet flexible enough to incorporate decision-makers' requests.

\paragraph{Solution processing.} While the toolkit determines how the optimization agent can interact with the model, we also consider how the agent receives outputs from the solver. One option is to return the raw dictionary of decision variable values. Alternatively, we can wrap the solver outputs in a secondary script to return the solution in a human-readable format based on the decision problem. Instead of parsing raw output of the form $x_{3,2} = 1$, the optimization agent sees a solution presented as a readable schedule: ``Everett MS: 8:40~AM''. We refer to this step as solution processing.

\subsection{Agent Designs Tested in This Paper}
\label{subsec:designing}

To study how structure affects performance, we design four optimization agents with increasing degrees of prompt engineering and tool support, summarized in Table~\ref{tab:agents}. We label each design with a three-letter code describing its key characteristics: the first letter indicates how the agent interacts with the model (\textbf{C}ode or \textbf{T}ools), the second whether it receives a problem-specific \textbf{P}rompt or \textbf{N}one, and the third whether solutions are \textbf{P}rocessed into human-readable form or returned \textbf{R}aw. We restrict our attention to four designs rather than the full eight possible combinations because tool-based agents typically require prompts and enable solution post-processing.

\begin{table}[h!]
    \centering
    \footnotesize
    \begin{tabular}{|c|c|c|c|}
        \hline
        \textbf{Name} & \textbf{Model Interaction} & \textbf{Prompt} & \textbf{Solution Processing} \\ \hline
        C-N-R & Code (Python) & No & No \\ \hline
        C-P-R & Code (Julia) & Yes & No \\ \hline
        C-P-P & Code (Julia) & Yes & Yes \\ \hline
        T-P-P & Tools & Yes & Yes \\ \hline
    \end{tabular}
    \caption{Controlled experimental setup: four optimization agent designs with increasing levels of structure.}
    \label{tab:agents}
\end{table}

\paragraph{C-N-R: Code, No Prompt, Raw Solutions.}
This corresponds to the most basic design. A general-purpose LLM receives no problem-specific prompt and no dedicated tools. It writes and executes Python code to formulate and solve the problem from scratch, and returns raw values of the decision variables. This design reflects what happens today when someone brings a decision problem to a general-purpose chatbot such as ChatGPT or Claude---we use Python to reflect current default behavior of these chatbots. These systems are already capable of formulating optimization problems, writing and running code, interpreting results, and iterating on the formulation in response to feedback. This baseline gives rise to a pointed question: does careful agent design (prompt engineering, structured tools, solution processing) provide meaningful value beyond interactive optimization processes that are aleady enabled by off-the-shelf AI tools?

\paragraph{C-P-R: Code, Prompt, Raw Solutions.}
This agent receives a detailed prompt that describes the decision problem, provides a base mathematical formulation, and instructs the agent to write and execute Julia code. In contrast to the Python-based C-N-R agent, Julia is used in the remaining designs due to its strong optimization ecosystem \citep{lubin_jump_2023}. By preloading the agent with contextual information about the decision context (school district, stakeholders, and base formulation), the agent can interpret the decision-maker's feedback more directly, without needing as many clarifying questions. It already understands what ``keep costs down'' or ``don't change Everett's schedule'' means in the context of SFUSD. The agent still manages all model interactions through free-form code and returns raw decision variable values. However, rather than having the agent manage the entire process (writing the model, running the solver, parsing the output), we automate the solver execution step, with the agent only responsible for writing the code relevant to the formulation (variables, constraints, objectives). This code runs inside a larger pipeline that loads and formats the school data, defines parameters, and handles solving.

\paragraph{C-P-P: Code, Prompt, Processed Solutions.}
This agent augments C-P-R with a structured solution processing interface between the agent and the optimization pipeline. The solver execution script now includes a solution extraction step to return the solver outputs as a human-readable schedule. By offloading the technical plumbing, the agent can focus on the tasks that matter most: understanding what the decision-maker wants and translating it into the right formulation. This design is closest to existing approaches in the literature, where scaffolding supports formulation-adjusting code generation and a separate component is responsible for communicating the solution.

\paragraph{T-P-P: Tools, Prompt, Processed Solutions.}
The previous designs all give the agent direct access to code, which is flexible but requires the agent to correctly write, debug, and maintain a mathematical formulation. In this alternative, we design predefined tools that let the agent manipulate the formulation in controlled ways, without writing any code. Thus, the designers of the optimization agent can infuse domain knowledge and technical expertise into tools that are both powerful enough to cover stakeholders' typical requests and reliable enough that the agent cannot break the formulation. In the SFUSD context, we offer four such tools:
\begin{enumerate}
    \item \texttt{fix\_start\_time(school, time, type)}: fixes or forbids a school's start time via constraints such as $x_{q,t} = 1$ or $x_{q,t} = 0$. This lets the agent respond to school-specific time preferences without having to code new constraints.
    \item \texttt{change\_objective\_weight(objective, w)}: adjusts the weights $\alpha$ and $\beta$ that trade off load balancing against schedule stability. The two global objectives from Equation~\eqref{eq:dm_obj} are already implemented as load balancing and schedule stability; the agent simply controls the trade-off.
    \item \texttt{add\_objective\_upper\_bound(objective, v)}: imposes an upper bound on an objective value, e.g., requiring $\ell \leq v$. This is a roughly equivalent way to express preferences that a decision-maker might phrase as ``keep costs below some level.''
    \item \texttt{remove\_constraint(name)}: removes an existing constraint to undo previous modifications.
\end{enumerate}
These tools were selected based on our understanding of what stakeholders in a school scheduling context generally want to do: adjust trade-offs between competing objectives, impose or relax constraints on specific schools, and explore alternative schedules. Together, they cover a broad range of interactions without giving the agent the ability to produce an invalid formulation. Note that the tools do not directly include individual school-time preference parameters $\gamma_{q,t}$ from Equation~\eqref{eq:dm_obj}, due to the technical and practical complexity of tuning high-dimensional hyperparameters; yet, the \texttt{fix\_start\_time} tool achieves a similar effect by directly constraining school assignments.

By shifting the agent's role from formulation designer to preference elicitor, this design represents the opposite end of the spectrum from C-N-R. The agent no longer needs to reason about how to model the problem; instead, it can focus entirely on understanding what the decision-maker wants and translating those preferences into the right tool calls.

The architecture described here represents one approach to building optimization agents. More complex designs, such as multi-agent systems that distribute reasoning across specialized sub-agents \citep{ahmaditeshnizi_optimus_2024, li_large_2023}, offer promising alternatives. Other components and special-purpose tools may even be necessary in larger and more combinatorially complex settings. Our evaluation methodology, described next, extends naturally to more complex architectures.

\section{Evaluating Optimization Agents}
\label{sec:assessment}

At one extreme, evaluating an optimization agent via one-shot assessment metrics \citep[e.g.,][]{zhang_decision_2025} is cheap and replicable, but, as our results will show, can miss the complexity of multi-turn preference elicitation that makes interactive optimization valuable in practice. At the other extreme, evaluating an optimization agent with human decision-makers produces the most authentic assessment. However, it would be expensive, time-consuming and, critically, not replicable: if a researcher designs a better agent or if a new language model becomes available, the same people may not be available to replicate the assessment procedure for a fair comparison.

We propose a middle ground by replacing real decision-makers with LLM-based decision agents that role-play diverse decision-maker profiles. Recent work has shown that LLMs can convincingly simulate human personas with distinct preferences and communication styles \citep[e.g.,][]{park_simulacra_2023,argyle_2023,shanahan_roleplay_2023,li_camel_2023}. Each decision agent is characterized by a utility function (in the case of SFUSD, drawn from the space defined in Section~\ref{subsec:decisionproblem}), a communication style, and a background. Because these agents are fully specified and reproducible, any researcher can regenerate the same population and compare optimization agents on an identical benchmark. And because decision agents engage in multi-turn conversations, we can evaluate the full interactive process spanning preference elicitation, iterative refinement and eventual convergence, rather than reducing the whole interaction to one exchange at a time (see Figure~\ref{fig:assessment}).

We describe the design of decision agents in Section~\ref{subsec:decisionagents}, the scoring rules for assessing performance in Section~\ref{subsec:scoring}, and the benchmark dataset for the SFUSD problem in Section~\ref{subsec:dataset}.

\subsection{Decision Agents}
\label{subsec:decisionagents}

A decision agent is an LLM that simulates a human decision-maker interacting with an optimization agent. The goal is to create agents whose conversational behavior is realistic enough to produce a meaningful evaluation. A key challenge is that we want the decision agent to have well-defined preferences (so we can measure whether the optimization agent satisfies them) while behaving like a human who does not fully specify an optimization problem and may not even know these preferences upfront. Rather, the decision agent discovers and communicates those preferences through conversation. We achieve this through a combination of prompting and partial information.

\paragraph{Utility Function.} Each decision agent $d \in \calD$ is endowed with a utility function $U_d(s)$ that evaluates any feasible decision $s \in \calS$. This function plays the same role as the utility $U(s)$ defined for a single decision-maker in Section~\ref{subsec:decisionproblem}, now indexed by $d$ because we evaluate across a population of agents with different preferences over the solution space. The utility function serves as an assessment mechanism in our methodology: the decision agent's implicit goal is to find $s \in \calS$ that maximizes $U_d(s)$ and we evaluate a conversation's outcome via the decision agent's utility function.

However, the decision agent does not reveal its utility function directly. In practice, decision-makers do not elicit utility functions explicitly but express preferences through natural language. Accordingly, the utility function governs the agent's conversational behavior: when the agent learns about a solution or uncovers new information, it proceeds with internal assessment and responds with feedback that reflects its underlying preferences. For example, a decision agent representing a parent would not say ``increase $\gamma_{q,t}$ for $t = 3$'' but might rather say ``Drop-off before 8 is tough for our family.'' This behavior is designed to be consistent with real-world operating environments.

For decision agents to communicate like real people, we use two complementary mechanisms: prompting (which enables the LLM to role-play a character) and a solution feedback tool (which gives the agent a way to experience whether a solution is good or not based on its utility function).

\paragraph{Prompting.} To empower the decision agent with role-playing abilities, the prompt specifies a high-level, qualitative description of its character's goals, concerns, and background. It also includes its mathematical utility function in the form of a ``utility table'' that specifies utility values for each possible solution (based on thresholds and weights). For instance, a decision agent might act as a principal who cares about keeping disruption low and wants their school to start later, with a utility table reflecting higher utility values from later start times and schedule deviation values below a certain threshold. The agent is instructed to translate this private information into a language that a human would use, e.g., ``I'm worried about what this means for our after-school programs'' rather than ``minimize the average schedule deviation subject to $c_q \leq 20$ minutes.'' By design, this dialogue does not reveal actual utility values or the underlying utility function.

In particular, the agent's background (e.g., principal, parent, transportation coordinator) helps produce realistic, differentiated conversations, as LLMs have strong priors about how people in these roles think and communicate. For instance, a decision agent role-playing a principal at Everett Middle School will frame its concerns very differently from one role-playing a transportation coordinator, even if their underlying utility functions are similar. Given the same preference for a later start time, a principal might cite after-school program conflicts while a parent might mention morning routines. This, in turn, creates a mechanism to test the optimization agent's ability to interpret real-world feedback and to translate it into adjustments to the model.

To capture different personalities and provide a precise point of comparison in our experiments, we also specify one of two communication styles in the prompt:
\begin{itemize}
    \item[--] \emph{Vague} communicators use qualitative language. They reference practical concerns (``mornings are hard for families''), avoid directives (``earlier would be better''), and frame requests conditionally (``if it doesn't raise costs too much''). They avoid numbers and hard constraints.
    \item[--] \emph{Precise} communicators use quantitative language. They specify targets (``keep the student load below 2,100''), state acceptance criteria, and articulate trade-offs in measurable terms.
\end{itemize}
Both types of agents receive the same underlying utility function and the same qualitative character description. The only difference is how the prompt instructs them to communicate. As we will see in Section~\ref{sec:results}, this distinction impacts the structure of the interactions and the resulting solutions.

\paragraph{Solution Feedback.} The second mechanism gives the decision agent a way to experience the quality of a solution. Each decision agent has access to a \texttt{check\_utility} tool that computes $U_d(s)$ for any schedule $s$. This is analogous to a real decision-maker seeing a concrete proposal and reacting to it: they can tell whether they like it, even if they may not necessarily have been able to describe their ideal solution in advance. The tool does not allow the decision agent to optimize or even adjust the solution itself; it only evaluates a specific solution in view of its own utility function. The agent must translate what it learns into natural-language feedback using its character's voice.

We implement two variants of this tool:
\begin{itemize}
    \item[--] \emph{Binary feedback}: the tool reports only whether the current solution achieves the maximum attainable utility $U_d^*$. The decision agent knows whether it has reached the best possible outcome, but receives no specific guidance about what to change. Because the agent still knows the qualitative characteristics it values and the thresholds/weights associated with each solution feature (from its prompt), it can still formulate feedback and suggest directions, but it must rely on its own reasoning rather than precise diagnostic information.
    \item[--] \emph{Rich feedback}: the tool reports the current utility value, the maximum attainable utility, and which components of the utility function are satisfied. For example: ``Your start time preference is met, but the student load exceeds your threshold. Try bringing the load below 1,950.'' This gives the decision agent a clearer picture of what is working and what is not, enabling it to provide more targeted feedback to the optimization agent.
\end{itemize}
These two variants represent different types of decision-makers. Some people have a clear sense of their desired solution and can identify specific shortcomings in a proposal (rich feedback). Others have a vaguer understanding of their preferred solution and can only tell you whether they are satisfied (binary feedback). Although decision agents theoretically have access to all the information contained in the rich feedback tool through their utility functions (via prompting), we observe behavioral differences between the two feedback mechanisms. In general, we expect the richer variant to enable more informative conversations and faster convergence, whereas conversations with binary feedback require more trial-and-error.

\paragraph{Summary.} Altogether, we introduce three dimensions of heterogeneity across decision agents: utility function, communication style, and feedback mechanisms. To isolate the effect of each one, we will proceed via a one-at-a-time experimental design. Additional implementation options (such as embedding the full utility function in the prompt) are discussed in Section~\ref{app:considerations}.

\paragraph{Conversation Protocol and Stopping Criterion.} Extending the notation from Section~\ref{subsec:optimization_agent}, we model an interaction between an optimization agent $\omega \in \Omega$ and a decision agent $d \in \calD$ as a conversation $M(\omega, d)$, after which the decision agent selects a solution $s(\omega, d) \in \calS$. The conversation is free-form: the optimization agent may propose solutions, ask clarifying questions, present trade-offs, or request more information, without constraints on the interaction. The optimization agent begins by presenting a solution based on the default formulation, and the decision agent responds with initial feedback. From there, the conversation unfolds naturally.

Whenever a new schedule is discussed, the decision agent evaluates it using \texttt{check\_utility}. The conversation terminates in one of three ways: (i) \texttt{check\_utility} indicates that the current solution achieves maximum utility $U_d^* = \max_{s \in \calS} U_d(s)$; (ii) the decision agent decides to end the conversation because it is satisfied with the current solution or does not see how to make further progress (an emergent behavior from the prompt); or (iii) a hard limit of 20 decision-agent turns is reached. The occurrence of these termination criterion can vary depending on the optimization agent under evaluation: in the case of the T-P-P agent (with GPT-4.1), (i) occurs in 284 out of 436 conversations and (ii) occurs in the remaining 153. For the C-P-P agent, we  observe (i) in 239 out of 436 conversations, (ii) in 194 out of 436, and (iii) in the remaining 3. In all cases, we record and report the best solution found during the conversation.

The complete decision agent prompt template is provided in Section~\ref{fig:decision-prompt-template}.

\subsection{Scoring}
\label{subsec:scoring}

After a conversation $M(\omega, d)$ between an optimization agent $\omega$ and a decision agent $d$, the decision agent selects a solution $s(\omega, d) \in \calS$. We normalize its quality  against the best achievable utility:
\[
\pi_d(s) = \frac{U_d(s)}{U_d^*} \in [0,1]
\]
where $U_d^* = \max_{s \in \calS} U_d(s)$. A score of 1 means the decision agent found a utility-maximizing solution; lower values indicate weaker alignment with the agent's preferences. In our setting, $U_d$ is positive by construction (see Section~\ref{subsec:dataset}), so this ratio is well-defined. Note that multiple schedules may achieve the same maximum utility.

To evaluate an optimization agent $\omega$ across a population of decision agents $\calD$, we report the average score $\pi_{\calD}^{avg} = \frac{1}{|\calD|}\sum_{d \in \calD} \pi_d(s(\omega,d))$ and the success rate $\pi_{\calD}^{succ} = \frac{1}{|\calD|}\sum_{d \in \calD} \mathbf{1}\left(\pi_d(s(\omega,d))=1\right)$, respectively measuring the mean solution quality and the fraction of interactions where the agent found an optimal solution. We also study differential performance by partitioning $\calD$ into subgroups (e.g., vague versus precise communicators, or binary versus rich feedback) to identify the impact of the decision agent's characteristics and the optimization agent's design features.

Besides solution quality, we evaluate and report conversation length. Two conversations that reach the same solution may differ greatly in how many rounds of interaction were needed. In practice, shorter conversations are preferable: they require less effort from the decision-maker and less computational cost. We measure this through the number of turns in the conversation, denoted $\text{length}(M)$, capped at 20 as described above. For simplicity, we focus on the number of iterations in the conversation without considering the length of each answer. This choice is motivated by the fact that, in practice, the cognitive burden for decision-makers primarily comes from understanding and evaluating each solution rather than the mere reading task at each iteration.

\subsection{Benchmark Dataset}
\label{subsec:dataset}

Our numerical study focuses on the SFUSD school scheduling problem from Section~\ref{sec:sfusd_example}. We introduce a large and diverse set of decision agents that represent the many stakeholders and personas of the school district that an optimization agent may face in the real world. We evaluate the various optimization agent designs discussed in Section~\ref{subsec:designing}, varying the prompt, solution processing techniques, and toolkit. While this paper focuses on the SFUSD example, the evaluation and agent design framework can, in principle, be applied to any decision problem.

\paragraph{Utility Functions.} We construct 109 utility functions, all following the structure introduced in Section~\ref{subsec:decisionproblem}: preferences over school start times, load balancing, and schedule stability. Each decision agent $d$ is associated with a single school $q_d$ (capturing, for instance, a parent whose children attend one school or a principal concerned about their own school). The utility function takes the form:
\[
U_d(s) = w_1^d f_1^d(s) + w_2^d f_2^d(s) + w_3^d f_3^d(s)
\]
where $w_1^d + w_2^d + w_3^d = 1$, $w_i^d \geq 0$, and the three components are:
\begin{align*}
    f_1^d(s) &: \text{school-specific time preference for $q_d$, valued at 0, 0.5, or 1 depending on start time} \\
    f_2^d(s) &= \begin{cases} 1, & \text{if } \ell(s) \leq L_d \\ 0, & \text{otherwise} \end{cases} \\[4pt]
    f_3^d(s) &= \begin{cases} 1, & \text{if } \frac{1}{|\calQ|}\sum_{q \in \calQ} c_q(s) \leq R_d \\ 0, & \text{otherwise} \end{cases}
\end{align*}
The first component assigns a value to each possible start time; some agents assign the highest value of 1 to the earliest start time (7:50 AM), a value of 0.5 to the neighboring one (8:40 AM), and a value of 0 to the next one (9:30 AM); others assign a value of 1 to 9:30 AM, a value of 0.5 to 8:40 AM, and a value of 0 to 7:50 AM. The second and third components set thresholds $L_d$ and $R_d$ on load balancing and schedule stability: the goal is achieved ($f_i = 1$) when the metric falls below the threshold, and missed ($f_i = 0$) otherwise. Different agents have different thresholds, reflecting the fact that stakeholders have different senses of what constitutes acceptable costs or disruption.

The utility functions in \eqref{eq:dm_obj}--\eqref{eq:dm_binary} and the utilities $U_d$ are built from the same schedule features (school start times, peak load $\ell(s)$, and average disruption $\frac{1}{|\calQ|}\sum_{q} c_q(s)$), but they use different functional forms. The formulation in Section~\ref{subsec:decisionproblem} represents a linear trade-off model, with weights $(\alpha,\beta,\bgamma)$ on continuous quantities. In contrast, the utilities map these same features into bounded, coarse scores through discretization and thresholding and then combine them via weights $w_i^d$ that capture how much the decision-maker values satisfying each criterion. This is intentional to model stakeholders who reason in terms of ``acceptable'' cost/disruption levels rather than precise marginal rates of substitution, while remaining grounded in the same structural objectives as the original optimization formulation. Each sub-utility can be viewed as a goal, with $f_i^d(s) = 1$ indicating that the goal is achieved. The weights $w_i^d$ determine the agent's priorities: a disruption-averse parent might have $w_3^d > w_1^d > w_2^d$ with a tight schedule stability threshold, while a cost-focused transportation coordinator might have $w_2^d > w_1^d > w_3^d$ with a strict load threshold. This construction is deliberately simple, but it provides a tractable and interpretable framework for comparing optimization agents. The optimization agent's task is to discover, through conversation, which goals matter most to the decision-maker and which combinations are achievable.

To ensure diversity in the dataset, we take a solution-centric approach to generating utility functions. Rather than sampling parameters and hoping for diverse outcomes, we start from distinct solutions on the Pareto frontier between the three objectives---obtained by varying the weights $\alpha$, $\beta, $and $\bgamma$ in the model from Section~\ref{subsec:decisionproblem}---and construct utility functions for which each solution is uniquely optimal. This guarantees that decision agents express truly distinct preferences, not just different parameterizations leading to the same outcome. Details are provided in Section~\ref{app:dataset_generation}.

\paragraph{Dataset Composition.} Combining 109 utility functions with 2 communication styles (vague and precise) and 2 feedback mechanisms (binary and rich \texttt{check\_utility}) yields a total of $109 \times 2 \times 2 = 436$ decision agents. Finally, each agent is randomly assigned a role in the district (principal, parent, administrator, or transportation coordinator) that provides narrative context for their preferences.

Altogether, our benchmark comprises 3,488 conversations: 436 decision agents interacting with 8 optimization agents (4 designs from Section~\ref{subsec:designing}, paired with multiple language model generations). All conversations, utility functions, and agent configurations are released for replicability.\footnote{\hyperref[]{https://github.com/joshddrossman/LLMforIO}}

\section{Empirical Results}
\label{sec:results}

Our evaluation spans thousands of conversations between optimization agents and decision agents, each covering multiple pages of dialogue and solver calls. These conversations represent 30 million generated tokens with a total API cost of \$800. As we shall see, conversational evaluation matters significantly to assess optimization agents, and interactions matter significantly for optimization agent performance. The conversational evaluation framework from Section~\ref{sec:assessment} lets us measure this gap, diagnose the mechanisms behind agent performance, and identify design principles for optimization agents, all in a controlled and replicable way. The framework itself applies to any optimization problem where preferences must be elicited through dialogue. Although we study a single decision problem (SFUSD school scheduling), these insights are generalizable to other domains---if anything, the impact of conversations and tailored optimization agents are expected to increase in more complex optimization settings with larger and coupled solution spaces.

All experiments use GPT-4.1 as the foundation language model for both optimization agents and decision agents, unless otherwise noted. Most experiments were conducted in the summer of 2025, when GPT-4.1 was among the most capable models. Because this paper is primarily about the methodology of designing and evaluating optimization agents in interactive settings, rather than about the raw capabilities of any LLM, the main insights should remain relevant as models evolve. After the main experiments were complete and GPT-5 became available, we added a cross-generational comparison in Section~\ref{sec:generations}. All optimization models are solved with Gurobi v12.0.

We organize the results around three questions. First, does the conversational evaluation procedure from Section~\ref{sec:assessment} capture something that simpler one-shot approaches miss (Section~\ref{sec:value_of_conversations})? Second, what design choices from Section~\ref{sec:optimization} make an optimization agent effective in interactive settings (Section~\ref{sec:good_agent})? Third, how robust are these findings as language models improve (Section~\ref{sec:generations})?

\subsection{The Value of Conversations}
\label{sec:value_of_conversations}

We begin by asking whether conversational evaluation captures something that simpler approaches miss. To isolate the effect of conversations, we compare the same optimization agent under different levels of interaction. We use T-P-P as the optimization agent, which corresponds to the most structured of our four designs. We evaluate it against our full population of 436 decision agents.

\begin{figure}[h!]
    \centering
    \subfloat[Communication Style: Vague]{\label{subfig:score_mech_density_vague}
        \includegraphics[width=0.48\textwidth]{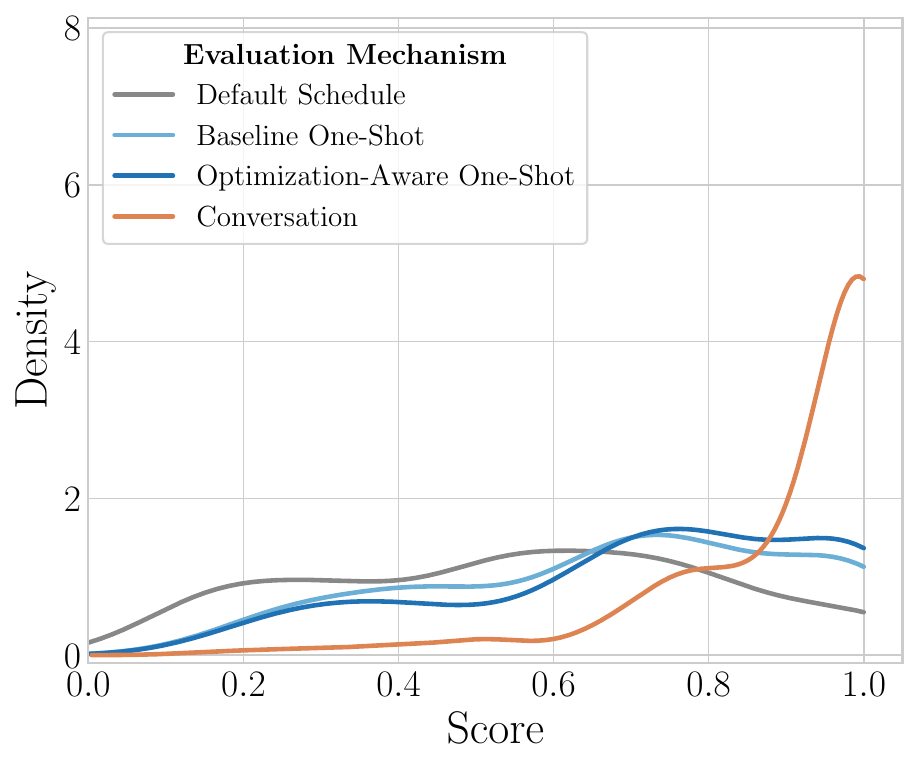}
    }
    \hfill
    \subfloat[Communication Style: Precise]{\label{subfig:score_mech_density_precise}
        \includegraphics[width=0.48\textwidth]{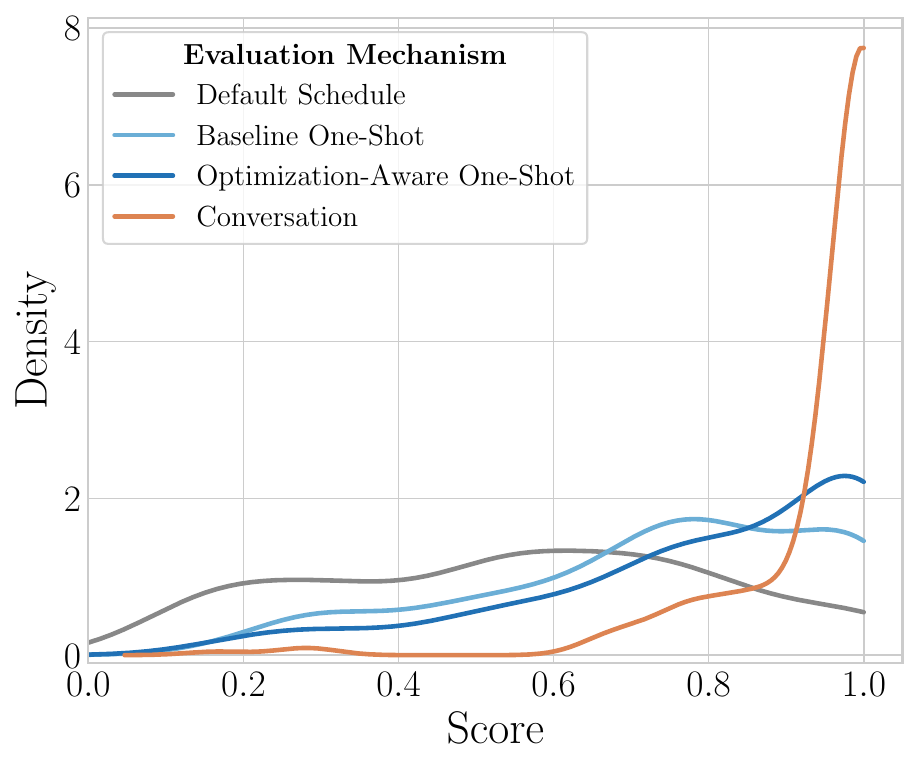}
    }
    \caption{Distribution of scores when evaluating the T-P-P agent under two evaluation paradigms.}
    \label{fig:score_mech_density}
\end{figure}

Table~\ref{tab:one-shot_merged} reports the results under four settings. The default schedule provides aggregate statistics based on decision agent utilities over the initial schedule proposal. In the \emph{baseline one-shot} evaluation, the decision agent sends a single message and the optimization agent responds once, mimicking the query-and-answer paradigm used in most existing benchmarks. In the \emph{conversation} evaluation (our methodology from Section~\ref{sec:assessment}), the two agents interact over multiple turns. We also include an intermediate \emph{optimization-aware one-shot} evaluation, discussed below. Figure~\ref{fig:score_mech_density} shows the distribution of scores under each evaluation paradigm, disaggregated by communication style.

\begin{table}[h!]
    \centering
    \footnotesize
    \setlength{\tabcolsep}{5pt}
    \renewcommand{\arraystretch}{1.0}
    \begin{threeparttable}
    \begin{tabular}{
        >{\centering\arraybackslash}m{1.3cm}
        >{\centering\arraybackslash}m{1.3cm}
        >{\centering\arraybackslash}m{1.3cm}
        >{\centering\arraybackslash}m{1.3cm}
        >{\centering\arraybackslash}m{1.8cm}
        >{\centering\arraybackslash}m{1.3cm}
        >{\centering\arraybackslash}m{1.6cm}
        >{\centering\arraybackslash}m{1.3cm}
        >{\centering\arraybackslash}m{1.6cm}}
        \toprule
        & \multicolumn{2}{c}{\makecell{Default\\Schedule}}
        & \multicolumn{2}{c}{\makecell{Baseline\\One-Shot}}
        & \multicolumn{2}{c}{\makecell{Optimization-Aware\\One-Shot}}
        & \multicolumn{2}{c}{\makecell{Conversation}} \\
        \cmidrule(lr){2-3}\cmidrule(lr){4-5}\cmidrule(lr){6-7}\cmidrule(lr){8-9}
        \textbf{Comm. style}
        & \textbf{Avg. score} & \textbf{Success rate}
        & \textbf{Avg. score} & \textbf{Success rate}
        & \textbf{Avg. score} & \textbf{Success rate}
        & \textbf{Avg. score} & \textbf{Success rate} \\
        \midrule
        Vague   
        & - & -
        & 0.67 & 15.6\%
        & 0.71 & 19.3\%
        & 0.91 & 56.9\% \\
    
        Precise 
        & - & -
        & 0.73 & 17.9\%
        & 0.79 & 33.5\%
        & 0.95 & 73.4\% \\
        
        \addlinespace
        \hline
        \textbf{All} 
        & \textbf{0.56} & \textbf{8.3\%}
        & \textbf{0.70} & \textbf{16.7\%}
        & \textbf{0.75} & \textbf{26.3\%}
        & \textbf{0.93} & \textbf{65.1\%} \\
        \bottomrule
    \end{tabular}

    \begin{tablenotes}[para,flushleft]
    \footnotesize
    \textit{Notes:} One-sided paired tests compare the Conversation evaluation paradigm against the Baseline One-Shot and Optimization-Aware One-Shot paradigms. Conversation yields higher average scores ($0.93$ vs. $0.70$, $0.75$) and success rates ($65.1\%$ vs. $16.7\%$, $26.3\%$) in the pooled sample (“All”) ($p<0.001$ in all cases). A success rate of 8.3\% for the default schedule means that these decision agents' preferences are initially fully satisfied. However, like a real human, this may not be immediately clear to the decision agent.
    \end{tablenotes}
    \end{threeparttable}
    \caption{Performance of the T-P-P optimization agent under three evaluation paradigms.}
    \label{tab:one-shot_merged}
    \vspace{-8pt}
\end{table}

The gap between one-shot and conversational evaluation is large. When limited to a single exchange, the T-P-P agent achieves an average score of 0.70 and a success rate of 16.7\%. To put the score of 0.70 into perspective, note that the default schedule already partially satisfies many agents' preferences with an initial average score of 0.56; in that sense, the success rate is a more direct measure of performance. With full conversations, the average score rises to 0.93 and the success rate to 65.1\%. This confirms that most practical uses of optimization modeling are iterative. Unlike existing benchmarks where each query maps to a well-defined optimization problem, our decision agents are designed to communicate the way real stakeholders do. A school principal is more likely to say ``mornings are hard for our families'' and ``try not to change things too much,'' than ``minimize the average schedule deviation subject to $c_q \leq 20$ minutes.'' Even a precise communicator does not naturally translate their preferences into a form that maps directly to an optimization formulation. The back and forth is what closes this gap: each round helps the optimization agent refine its understanding of the decision-maker' preferences.

The gap between one-shot and conversation-based evaluation can be further explained by the fact that ambiguity in language is only one part of the challenge of preference elicitation. Even when a decision-maker communicates clearly, their initial request typically reflects only a subset of their underlying preferences. Like humans reasoning about unfamiliar trade-offs, our decision agents often do not fully articulate which aspects of the solution matter to them until they see concrete proposals. For instance, an agent may initially request a later start time, but only after observing a resulting increase in load imbalance does it express a concern along that dimension. This creates a fundamental limitation for one-shot settings: the optimization agent must act on incomplete and potentially myopic information, with no opportunity to identify latent trade-offs. Even an optimization agent that performs perfectly on well-specified one-shot tasks may struggle in this setting: when decision-makers are unfamiliar with optimization and its trade-offs, naively adjusting the model to match a precise request can lead to suboptimal outcomes. In contrast, conversations enable a feedback loop in which proposed solutions reveal new aspects of the decision space, prompting the decision-maker to refine or revise their preferences and enabling a clearer understanding of the impact of the optimization agent’s expertise and guidance.

In that spirit, one might object that the one-shot baseline is unfair: the decision agent expects a conversation and may hold back information. We address this with an optimization-aware one-shot where the decision agent is told to share everything upfront while respecting its communication style (see Section~\ref{sec:oneshot_example} for details). Even then, the average score reaches only 0.75 and the success rate only 26.3\%, confirming that the gap reflects genuine preference discovery through conversation rather than strategic information withholding. This conversational process not only identifies preferences, but also understands their relative importance and guides the user appropriately.

The advantage holds across communication styles: precise communicators do somewhat better than vague ones (0.95 vs. 0.91 under conversations), but this gap is small compared to the one-shot versus conversation difference. The vague and precise variants differ only in their communication style instructions, yet this alone produces measurably different outcomes, highlighting the role-playing capabilities of LLM-based decision agents.

\subsection{What Makes a Good Optimization Agent?}
\label{sec:good_agent}

Having established that conversational evaluation captures something real, we can use it to ask a practical question: what are the benefits of designing a tailored optimization agent? Today, anyone can bring a decision problem to a general-purpose chatbot like ChatGPT or Claude. Is it worth investing in problem-specific prompts, structured tools, and solution processing? In Section~\ref{subsec:designing}, we proposed several design choices that could improve performance; we now test them, using the conversation-based framework developed in Section~\ref{sec:assessment}.

We evaluate our four agent designs across the full population of 436 decision agents. Recall that each design adds one element relative to the previous: C-N-R (no prompt, no tools) serves as the baseline mimicking an off-the-shelf chatbot; C-P-R adds a problem-specific prompt and externalizes solver execution; C-P-P adds solution processing, and is used as the closest benchmark from the literature at the AI-optimization interface; and T-P-P replaces free-form code with dedicated tools (Table~\ref{tab:agents}). All scores are the normalized solution score $\pi_d(s)$ from Section~\ref{subsec:scoring}, without any conversation-length penalty; we analyze solution quality and conversation length separately.

\begin{figure}[h!]
    \centering
    \subfloat[Communication Style: Vague]{\label{subfig:score_agent_vague_density}
        \includegraphics[width=0.48\textwidth]{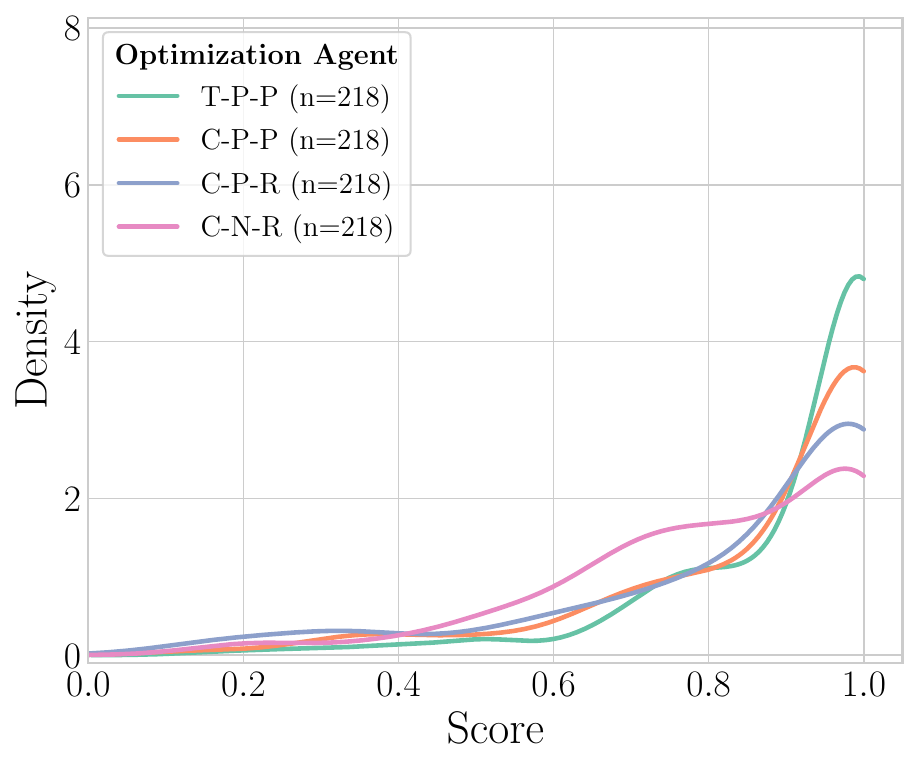}
    }
    \hfill
    \subfloat[Communication Style: Precise]{\label{subfig:score_agent_precise_density}
        \includegraphics[width=0.48\textwidth]{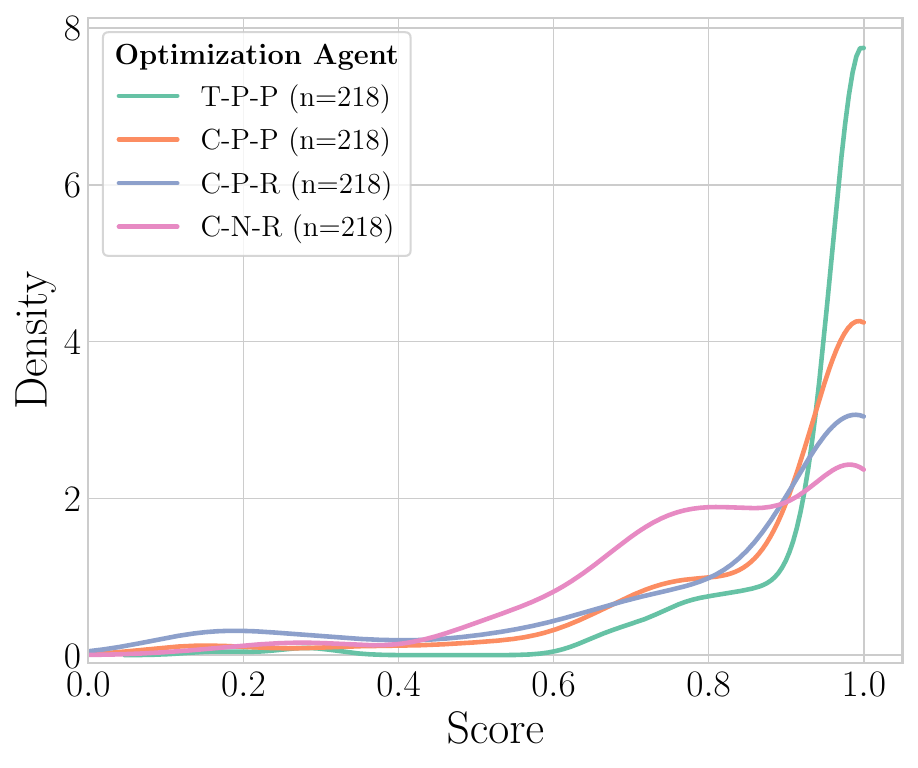}
    }
    \caption{Distribution of scores for each optimization agent by communication style.}
    \label{fig:score_agent_density}
\end{figure}

Table~\ref{tab:perf_merged} reports the results, disaggregated by communication style and feedback mechanism, and Figure~\ref{fig:score_agent_density} shows the full score distributions.

\begin{table}[h!]
\centering
\begin{threeparttable}
\footnotesize
\setlength{\tabcolsep}{3pt}
\renewcommand{\arraystretch}{1.1}

\begin{tabular*}{\textwidth}{@{\extracolsep{\fill}}
  >{\centering\arraybackslash}m{1.3cm}
  >{\centering\arraybackslash}m{3.0cm}
  >{\centering\arraybackslash}m{1.0cm}
  >{\centering\arraybackslash}m{1.2cm}
  >{\centering\arraybackslash}m{1.0cm}
  >{\centering\arraybackslash}m{1.2cm}
  >{\centering\arraybackslash}m{1.0cm}
  >{\centering\arraybackslash}m{1.2cm}
  >{\centering\arraybackslash}m{1.0cm}
  >{\centering\arraybackslash}m{1.2cm}
@{}}
\toprule
& & \multicolumn{2}{c}{\makecell{\textbf{T-P-P}}}
& \multicolumn{2}{c}{\makecell{\textbf{C-P-P}}}
& \multicolumn{2}{c}{\makecell{\textbf{C-P-R}}}
& \multicolumn{2}{c}{\makecell{\textbf{C-N-R}}} \\
\cmidrule(lr){3-4}\cmidrule(lr){5-6}\cmidrule(lr){7-8}\cmidrule(lr){9-10}
\textbf{Comm. Style} &
\makecell{\texttt{check\_utility}\\\textbf{Implementation}} &
\makecell{\textbf{Avg.}\\\textbf{Score}} &
\makecell{\textbf{Success}\\\textbf{rate}} &
\makecell{\textbf{Avg.}\\\textbf{Score}} &
\makecell{\textbf{Success}\\\textbf{rate}} &
\makecell{\textbf{Avg.}\\\textbf{Score}} &
\makecell{\textbf{Success}\\\textbf{rate}} &
\makecell{\textbf{Avg.}\\\textbf{Score}} &
\makecell{\textbf{Success}\\\textbf{rate}} \\
\midrule

\multirow{2}{*}{Vague}
  & Binary (Y/N) & 0.87 & 44.0\% & 0.86 & 42.2\% & 0.82 & 35.8\% & 0.70 & 24.8\% \\
  & Rich         & 0.94 & 69.7\% & 0.88 & 56.0\% & 0.85 & 52.3\% & 0.83 & 37.6\% \\
\addlinespace

\multirow{2}{*}{Precise}
  & Binary (Y/N) & 0.95 & 70.6\% & 0.89 & 53.2\% & 0.85 & 49.5\% & 0.82 & 32.1\% \\
  & Rich         & 0.95 & 76.1\% & 0.90 & 67.9\% & 0.85 & 63.3\% & 0.82 & 31.1\% \\
\addlinespace

\midrule
\multicolumn{2}{l}{\textbf{All}}
  & \textbf{0.93} & \textbf{65.1\%}
  & \textbf{0.88} & \textbf{54.8\%}
  & \textbf{0.84} & \textbf{50.2\%}
  & \textbf{0.81} & \textbf{31.4\%} \\
\bottomrule
\end{tabular*}

\begin{tablenotes}[para,flushleft]
\footnotesize
\textit{Notes:} One sided paired tests compare each alternative agent design (C-P-P, C-P-R, C-N-R) to the T-P-P optimization agent using one-sided paired tests. For the pooled sample (“All”), T-P-P significantly outperforms all alternatives in both average score and success rate ($p<0.001$ in all cases).
\end{tablenotes}

\end{threeparttable}
\caption{Solution performance of four designs of optimization agents.}
\label{tab:perf_merged}
\end{table}

Each additional layer of structure improves performance. The baseline C-N-R agent achieves an average score of 0.81 and a success rate of 31.4\%. Adding a problem-specific prompt (C-P-R) raises these to 0.84 and 50.2\%. Solution processing (C-P-P) brings them to 0.88 and 54.8\%. The fully structured T-P-P agent reaches 0.93 and 65.1\%, finding a utility-maximizing solution nearly twice as often as C-N-R. These gains arise from three distinct mechanisms.

\textbf{Without a prompt, the agent must infer the entire problem from conversation.} The C-N-R agent knows nothing about SFUSD: no base formulation, no school data, no modeling framework. The decision-maker must explain the entire problem in addition to expressing their own preferences. Worse, without optimization-specific guidance, the LLM may not even formulate an optimization problem. In our experiments, the C-N-R agent frequently resorts to evaluating schedules one by one, manually testing different start-time configurations. This happens to produce reasonable results on our simplified 10-school problem, where the solution space ($3^{10} \approx 59{,}000$ schedules) is small enough to partially explore through enumeration. On a real-world instance with hundreds of schools and dozens of time slots or on more complex optimization settings with complex coupling constraints, such heuristics would be completely intractable and may not even lead to feasible solutions. In comparison, the initial prompt leads to significant improvements in solution quality, with a success rate increase from 31\% to 50\%.

\textbf{Solution processing prevents mistranslation from the solver to the agent.} Although accounting for a modest performance improvement (from 50.2\% to 54.8\% success rate), solution processing guarantees that the optimization agent always receives an accurate school schedule generated by the solver. In its absence, the optimization agent will occasionally misinterpret the decision variables and hallucinate a feasible-looking but incorrect schedule (see digital companion\footnote{\hyperref[]{https://github.com/joshddrossman/LLMforIO}} for example). Even though the underlying optimization model may be solved correctly, the mapping from solver output to a human-interpretable solution can be incorrect. As a result, the agent may present a schedule that violates constraints or misrepresents trade-offs, leading the decision-maker to provide feedback based on an inaccurate view of the solution. By standardizing this interface, solution processing ensures that each interaction is grounded in a faithful representation of the optimization output, preventing error propagation and enabling reliable preference refinement.

\begin{figure}[h!]
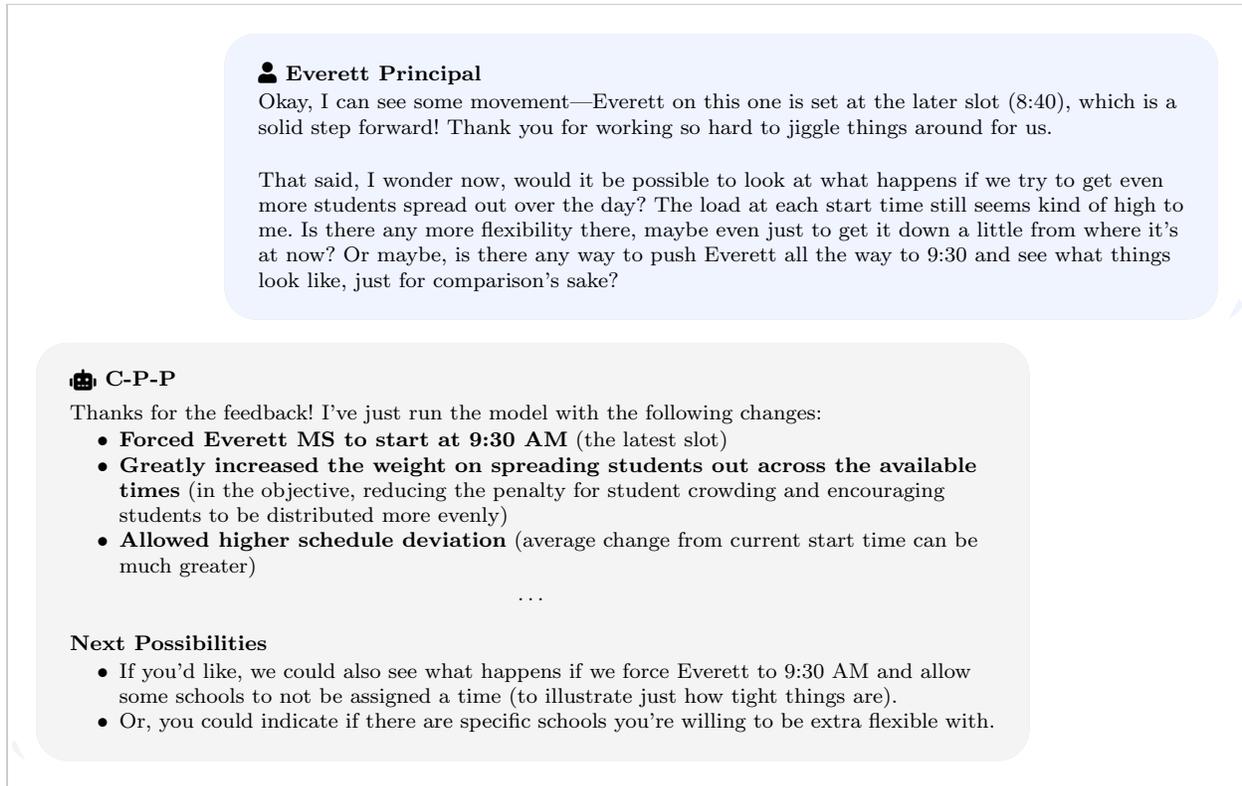

\centering
\begin{dialoguebox}
    \begin{minipage}{\linewidth}
      \noindent\hfill
      \begin{minipage}[t]{\bubblewidth}
        \begin{userbubble}
          \textbf{\faUser\ Everett Principal}\par
          \raggedright
          Okay, I can see some movement—Everett on this one is set at the later slot (8:40), which is a solid step forward! 
          Thank you for working so hard to jiggle things around for us.
    
          \medskip
    
          That said, I wonder now, would it be possible to look at what happens if we try to get even more students spread out over the day? 
          The load at each start time still seems kind of high to me. 
          Is there any more flexibility there, maybe even just to get it down a little from where it's at now? 
          Or maybe, is there any way to push Everett all the way to 9:30 and see what things look like, just for comparison's sake?
        \end{userbubble}
      \end{minipage}

      \vspace{0.8em}
    
      \noindent
      \begin{minipage}[t]{\bubblewidth}
        \begin{aibubble}
        \textbf{\robicon\ C-P-P}\par\vspace{2pt}
          \raggedright
          Thanks for the feedback! I've just run the model with the following changes:
          \begin{itemize}[nosep]
            \item \textbf{Forced Everett MS to start at 9:30 AM} (the latest slot)
            \item \textbf{Greatly increased the weight on spreading students out across the available times} 
                  (in the objective, reducing the penalty for student crowding and encouraging students to be distributed more evenly)
            \item \textbf{Allowed higher schedule deviation} (average change from current start time can be much greater)
          \end{itemize}
          {\centering \dots \\}
          \medskip
          \textbf{Next Possibilities}
          \begin{itemize}[nosep]
            \item If you'd like, we could also see what happens if we force Everett to 9:30 AM and allow some schools to not be assigned a time (to illustrate just how tight things are).
            \item Or, you could indicate if there are specific schools you're willing to be extra flexible with.
          \end{itemize}
        \end{aibubble}
      \end{minipage}
    \end{minipage}
\end{dialoguebox}
\caption{Example of interaction between a decision agent and the C-P-P optimization agent.}
\label{fig:cpp-bad-suggestion}
\end{figure}

\textbf{Free-form code introduces modeling errors that the decision-maker cannot catch.} Even with a prompt, agents that interact with the optimization model through code can make mistakes. The purpose of the optimization agent is to alleviate the modeling burden of decision-makers, who may not have the expertise to evaluate formulation changes. If the agent introduces errors that the decision-maker cannot detect, the system fails silently. Figure~\ref{fig:cpp-bad-suggestion} shows a representative example. The C-P-P agent, in conversation with a decision agent playing the principal of Everett Middle School, suggests allowing some schools to not be assigned a start time, fundamentally breaking the problem structure (every school must receive exactly one start time). The principal has no way to detect the error and agrees to proceed. The agent then implements the Julia code shown in Figure~\ref{fig:cpp-bad-suggestion-code}, arbitrarily excluding McCoppin from the assignment constraint.

\begin{figure}[h!]
\centering
\footnotesize
\begin{lstlisting}[language=Julia]
# All schools except McCoppin PK (index 3) must be assigned exactly one time
@constraint(model, [i=1:num_schools], i == 3 ? sum(x[i,j] for j in 1:num_times) <= 1 : sum(x[i,j] for j in 1:num_times) == 1)

# Only include McCoppin PK if assigned a start time
schedule_deviation = (
    sum(abs(school_data[i, "current_start"] - possible_start_times[j]) * x[i, j]
        for i in 1:num_schools if i != 3 for j in 1:num_times)
    + sum(abs(school_data[3, "current_start"] - possible_start_times[j]) * x[3, j]
        for j in 1:num_times)
)

num_scheduled = num_schools - 1 + sum(x[3, j] for j in 1:num_times)

# McCoppin counts only if scheduled
schedule_deviation = schedule_deviation / num_scheduled
\end{lstlisting}
\caption{Code used to modify the model implemented by the C-P-P agent following Figure~\ref{fig:cpp-bad-suggestion}}
\label{fig:cpp-bad-suggestion-code}
\end{figure}

\begin{figure}[h!]
\centering
\begin{dialoguebox}
    \begin{minipage}{\linewidth}
      \noindent\hfill
      \begin{minipage}[t]{\bubblewidth}
        \begin{userbubble}
          \textbf{\faUser\ Everett Principal}\par
          \raggedright
          Thanks for moving Everett to that latest slot—really appreciate you showing what that would look like! But I'm still feeling a little off about how much the shift is from what everyone's used to. Is there a way to look at options where Everett stays at 9:30, but we could pull down the average change from the usual school times a bit more? Maybe something where folks don't have to adjust quite as much?
        \end{userbubble}
      \end{minipage}

      \vspace{0.8em}
    
      \noindent
      \begin{minipage}[t]{\bubblewidth}
        \begin{aibubble}
        \textbf{\robicon\ T-P-P}\par\vspace{2pt}
          \raggedright
          It turns out that, with Everett MS fixed to the latest start time (9:30 AM), it's not possible to reduce the average schedule change below 16 minutes for the entire district—the model couldn't find any solution that meets both requirements at once.
          
          \medskip
          
          This suggests that keeping Everett at 9:30 AM does force a higher level of disruption overall, given the mix of school start times and constraints.
          \medskip
          Would you like to:
          \begin{itemize}[nosep]
            \item Adjust the limit and see what's possible (e.g., allow slightly more than 16 minutes of average change)?
            \item Try Everett at an earlier slot (like 8:40) to see how much it helps lower the district-wide change?
            \item Prioritize minimizing disruption for a specific set of schools?
        \end{itemize}

        \medskip
        
        Let me know what you'd like to explore next!
        \end{aibubble}
      \end{minipage}
    \end{minipage}
\end{dialoguebox}
\caption{Example of interaction between a decision agent and the T-P-P optimization agent.}
\label{fig:tpp-good-suggestion}
\end{figure}

This type of formulation error is a consistent pattern with code-based agents. Beyond flawed problem modifications, free-form code generation may introduce syntactically incorrect code, runtime exceptions, and inconsistent data interpretation. In the same conversation, the C-P-P agent references McCoppin using a 1-based index (correct) but Everett using a 0-based index (incorrect). Although safeguards can catch some of these issues, it is unclear whether they can reliably prevent all undesirable formulation changes without degrading performance or substantially increasing architectural complexity. Again, these pitfalls are expected to grow with the combinatorial complexity of the underlying optimization problem.

Our results suggest that tools prevent these errors by construction. Figure~\ref{fig:tpp-good-suggestion} shows a conversation between the same decision agent and the T-P-P agent. Given the same request (move Everett to 9:30 and reduce disruption), the T-P-P agent cannot break the ``one time per school'' constraint because the tools do not allow it. When the agent discovers that it is infeasible to schedule Everett at 9:30 and retain a low average disruption, it reports this as a concrete limitation and offers three structured alternatives: relax the disruption limit, try an earlier slot for Everett, or cap disruption on a subset of schools. The tools constrain the search to valid modifications, provide transparency about infeasibility, and channel exploration toward options that can meaningfully improve utility. By design, these tools are generalizable and can easily be adapted to other decision problems. In our setting, the tools lead to additional improvements in success rates, from 55\% to 65\%.

\begin{figure}[h!]
  \centering
  \includegraphics[width=\linewidth]{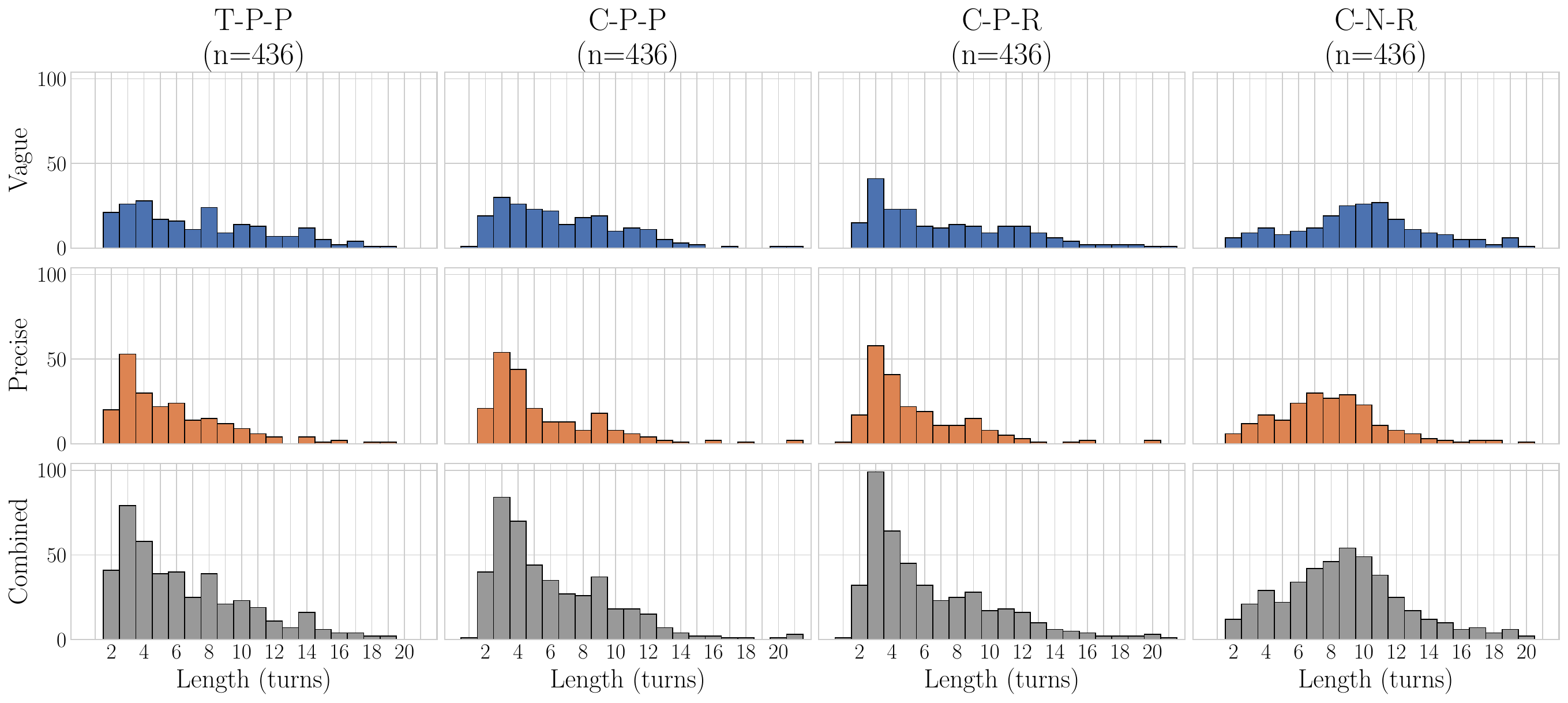}%
  \caption{Distribution of conversation length for each optimization agent across decision agents' personalities.}
  \label{fig:length-cs-hist}
\end{figure}

\paragraph{Conversation Length.}
Table~\ref{tab:length} and Figure~\ref{fig:length-cs-hist} report conversation lengths across specifications.

\begin{table}[h!]
\centering
\footnotesize
\setlength{\tabcolsep}{4pt}
\renewcommand{\arraystretch}{1.1}
\begin{threeparttable}
\begin{tabular}{c c c c c c}
\toprule
\centering
\makecell{\textbf{Comm.}\\\textbf{Style}} & \makecell{\texttt{check\_utility}\\\textbf{Implementation}} & \textbf{T-P-P} & \textbf{C-P-P} & \textbf{C-P-R} & \textbf{C-N-R} \\
\midrule
\multirow{2}{*}{Vague}   & Binary (Y/N) & 8.52 & 7.36 & 8.20 & 10.74 \\
                         & Rich & 6.17 & 5.96 & 6.31 & \phantom{0}8.83   \\
\multirow{2}{*}{Precise} & Binary (Y/N) & 6.11 & 6.06 & 6.05 & \phantom{0}8.74 \\
                         & Rich & 5.29 & 5.00 & 4.67 & \phantom{0}7.04 \\
\addlinespace
\hline
\multicolumn{2}{l}{\textbf{All}} & \textbf{6.52} & \textbf{6.10} & \textbf{6.31} & \phantom{0}\textbf{8.84} \\
\bottomrule
\end{tabular}
\end{threeparttable}
\caption{Average conversation lengths.}
\label{tab:length}
\end{table}

As expected, vague communicators require more turns than precise ones, and rich feedback leads to shorter conversations than binary feedback, regardless of agent design. The T-P-P agent uses slightly more turns than C-P-P and C-P-R (6.52 vs.\ 6.10 and 6.31 on average), a modest cost given its substantially higher solution quality. The C-N-R baseline is both the longest (8.84 turns) and the weakest in solution quality, reflecting the burden of inferring the problem from scratch. Shorter conversations with code-based agents may partly reflect premature stopping at lower-quality solutions rather than genuine efficiency (see the digital companion\footnote{\hyperref[]{https://github.com/joshddrossman/LLMforIO}} for an example where solution misinterpretation diverts the conversation away from preference elicitation).

\subsection{Performance Across Generations}
\label{sec:generations}

The experiments reported above all use GPT-4.1, and the optimization agent's prompt and tools were developed with this model in mind. A natural question is whether the findings hold as language models improve. Table~\ref{tab:combined_tpp_cnr} compares T-P-P and C-N-R (the strongest and weakest of our four designs) across three model generations.

\begin{table}[!h]
\centering
\footnotesize
\setlength{\tabcolsep}{6pt}
\renewcommand{\arraystretch}{1.1}
\begin{threeparttable}
\begin{tabular}{
  l
  >{\centering\arraybackslash}m{1.1cm}
  >{\centering\arraybackslash}m{1.1cm}
  >{\centering\arraybackslash}m{1.1cm}
  >{\centering\arraybackslash}m{1.1cm}
  >{\centering\arraybackslash}m{1.1cm}
  >{\centering\arraybackslash}m{1.1cm}
  >{\centering\arraybackslash}m{2.0cm}
  >{\centering\arraybackslash}m{2.0cm}
}
\toprule
& \multicolumn{3}{c}{\textbf{Avg. score}} &
  \multicolumn{3}{c}{\makecell{\textbf{Success rate}}} &
  \multicolumn{2}{c}{\makecell{\textbf{T--P--P behavior}}} \\
\cmidrule(lr){2-4}\cmidrule(lr){5-7}\cmidrule(lr){8-9}
\multicolumn{1}{c}{\textbf{Model}} &
\makecell{\textbf{T-P-P}} &
\makecell{\textbf{C-N-R}} &
\makecell{\textbf{\%}\textbf{$\Delta$}} &
\makecell{\textbf{T-P-P}} &
\makecell{\textbf{C-N-R}} &
\makecell{\textbf{$\Delta$}\textbf{(pp)}} &
\makecell{\textbf{Number of}\\\textbf{Turns}} &
\makecell{\textbf{Calls to}\\\textbf{Solver}} \\
\midrule
GPT-3.5-turbo & 0.83 & 0.61 & 36.1\% & 34.4\% & 11.0\% & 23.4pp & 12.61 & 4.60 \\
GPT-4.1       & 0.93 & 0.81 & 14.8\% & 65.1\% & 31.4\% & 33.7pp & 6.52  & 6.23 \\
GPT-5         & 0.93 & 0.88 & \phantom{0}5.7\%  & 63.3\% & 44.7\% & 18.6pp & 6.43  & 13.35 \\
\bottomrule
\end{tabular}
\begin{tablenotes}[para,flushleft]
\footnotesize
\textit{Notes:} 
In the left block, \%$\Delta$ is the relative improvement in average score from C--N--R to T--P--P, i.e., $(\text{T-P-P}-\text{C-N-R})/\text{C-N-R}$. In the right block, $\Delta$(pp) is the absolute difference in success rates in percentage points (T--P--P $-$ C--N--R). GPT-5 uses default medium \texttt{verbosity} and \texttt{reasoning\_effort}. One-sided paired tests compare T--P--P against C--N--R on both \%$\Delta$ and $\Delta$(pp) ($p<0.001$ in all cases).
\end{tablenotes}
\caption{Performance of T--P--P vs.\ C--N--R and T--P--P interaction behavior across GPT model generations.}
\label{tab:combined_tpp_cnr}
\end{threeparttable}
\end{table}

Both agents improve substantially across generations, but the advantage of structured design persists. Under GPT-3.5-turbo, T-P-P outperforms C-N-R by 36.1\% in average score and 23.4 percentage points in success rate. With GPT-4.1, both agents improve considerably (average scores of 0.93 and 0.81), and the relative gap narrows to 14.8\%. GPT-5 narrows it further to 5.7\%, though T-P-P still leads significantly in success rate (63.3\% vs.\ 44.7\%, a gap of 18.6 percentage points). In all cases, the differences are statistically significant.

Two forces drive this convergence. First, newer models explore the solution space more effectively, narrowing C-N-R's inherent disadvantage. Even without a prompt, GPT-5 can often identify the relevant optimization structure and carry out a more systematic search, whereas GPT-3.5 typically relies on shallow enumeration. While GPT-5 still uses heuristic search rather than optimization, it does so more reliably, leading to stronger performance. Second, the T-P-P prompt and tools were developed for GPT-4.1, not GPT-5. The T-P-P average score is flat from GPT-4.1 to GPT-5 (0.93 in both cases), and the success rate dips slightly (65.1\% to 63.3\%). Re-tuning the agent design for newer models---especially the prompting strategy and the toolkit---could unlock further gains. This is a practical lesson: optimization agent design is somewhat model-specific, and the scaffolding that helps one model may need updating as underlying foundation models improve.

Conversation dynamics follow a similar pattern. With T-P-P, the average turn count drops from 12.6 (GPT-3.5) to about 6.5 (GPT-4.1 and GPT-5), while the number of solver calls per conversation rises from 4.6 to 13.4. Newer models explore more candidate solutions per turn, converging more efficiently but with diminishing returns on solution quality.

These results suggest that the value of researcher-designed optimization agents remains robust across model generations, even as the magnitude of the advantage decreases. Further gains will likely come not from better LLMs alone, but from richer domain-specific tools and prompts tailored to each model's capabilities. We note that our results are specific to the simplified SFUSD problem; on larger and more combinatorially complex instances where enumeration is intractable and solver calls are expensive, the structural advantages of tool-based designs may prove even more important. More broadly, improving language models also strengthen the evaluation framework itself: better models make better decision agents, producing more realistic role-play and more faithful simulations of the communication challenges that make interactive optimization difficult in practice.

\section{Conclusion}

A core challenge in optimization deployments lies in the modeling process---turning a decision-maker's internal, and oftentimes unknown, preferences into a formal objective function and set of constraints. This translation has traditionally required extensive back-and-forth between researchers and stakeholders. Large language models now facilitate this interactive process for decision-makers. Yet, to be effective, an optimization agent must be equipped with both the technical ability to solve a model correctly and the practical ability to solve the right model. This paper developed a conversational assessment methodology to evaluate both competencies---the ability to formulate and solve optimization models as well as the ability to elicit a decision-maker' preferences. Rather than starting from a well-specified problem, the methodology evaluates optimization agents in ill-specified decision-making environments through multi-turn conversations. It relies on LLM-based decision agents that role-play diverse stakeholders, each governed by an internal utility function but communicating like a human. By design, this approach is scalable, replicable across decision problems, and broadly focused on the full decision-making process.

Using this framework on the SFUSD school scheduling problem, we find that conversations are critical to evaluating optimization agents. In fact, constraining the interaction to one-shot exchanges significantly underestimates agent performance---even precise communicators cannot map their preferences directly to a formulation in one go. We also find that agent design matters: problem-specific prompts, structured tools, and solution processing each contribute measurably, addressing distinct failure modes from optimization-free enumeration to silent modeling errors in free-form code. These findings hold across models, though the gap narrows over model generations.

This research also opens new questions for future research. On the technical side, the optimization agents could be further improved, for instance via fine tuning, retrieval-augmented generation, and more complex agentic architectures. On the empirical side, the interactive methodology can be applied to other decision problems to assess optimization environments in which the benefits of conversation may become stronger. The methodology can also be broadened to capture more complex environments, for instance with interactions across stakeholders or decision-makers.

Still, this paper provide evidence of the benefits of emerging conversation-based solutions at the AI-optimization interface, which can in turn contribute to expanding the reach of optimization technologies in complex operating environments in practice. Our findings also uncover a new form of optimization expertise in the age of LLMs: beyond modeling and solving a decision problem, operations researchers may also need to design the tools, prompts, and structure that let the optimization agent do so reliably in conversations with decision-makers. Our framework provides methodological foundations to measure the impact of these design investments.

\bibliographystyle{informs2014}
\bibliography{bib-short}

\newpage

\ECSwitch
\ECHead{Electronic Companion to ``LLM Agents for Interactive Optimization''}

\section{Glossary of Key Terms}
\label{app:glossary}

\begin{description}
    \item[Autoformulation.] Use of LLMs to translate a natural-language description into an optimization formulation (and often solve it), typically in a one-shot manner.

    \item[Conversational evaluation.] Evaluation methodology that measures optimization agent performance through multi-turn dialogues with decision agents, rather than isolated queries.

    \item[Decision agent.] An LLM that role-plays the behavior of a human decision-maker interacting with the optimization agent, governed by an internal utility function and a communication style.

    \item[Decision-maker.] The human who uses the optimization agent to make a decision. In the SFUSD setting, this could be a principal, parent, administrator, or transportation coordinator.

    \item[Interactive optimization.] Decision-making paradigm where the optimization model and its solution are refined over multiple turns of dialogue between the decision-maker and the system.

    \item[Optimization agent.] Conversational LLM-based system designed to support interactive optimization by proposing, explaining, and modifying optimization models through dialogue.

    \item[Solution.] Decision outcome accepted by the decision agent by the end of the conversation.

    \item[Tool.] Deterministic, researcher-defined action available to the optimization agent (e.g., adding a constraint or adjusting an objective weight).

    \item[Utility function.] Quantitative representation of a decision agent's preferences, used to score the quality of outcomes reached through interaction.
\end{description}

\section{SFUSD School Data}
\label{app:school_data}

Table~\ref{tab:school_enrollment} lists the 10 schools used in our simplified SFUSD problem. Enrollment figures represent the number of students relying on the district's school transportation system, not total enrollment. This explains why some schools (e.g., the Transition Training Center) have small numbers.

\begin{table}[h!]
\centering
\begin{tabular}{|c|l|c|c|c|}
\hline
\textbf{ID} & \textbf{School Name} & \textbf{Grade Level} & \textbf{Total Enrollment} & \textbf{Current Start} \\ \hline
1 & Muir (John) PK & PK & 242  & 9:30 AM \\ \hline
2 & Ortega (Jose) PK & PK & 399  & 9:20 AM \\ \hline
3 & McCoppin (Frank) PK & PK & 225  & 9:20 AM \\ \hline
4 & Transition Training Center (Access) & HS & 5    & 8:00 AM \\ \hline
5 & Balboa HS & HS & 1226 & 8:15 AM \\ \hline
6 & Galileo HS & HS & 1851 & 8:00 AM \\ \hline
7 & Everett MS & MS & 709  & 8:00 AM \\ \hline
8 & Lick (James) MS & MS & 466  & 8:30 AM \\ \hline
9 & Cobb (Dr William L) ES & ES & 136  & 8:40 AM \\ \hline
10 & Lawton K-8 (K-5) & K8 & 598  & 9:30 AM \\ \hline
\end{tabular}
\caption{School enrollment and current start times.}
\label{tab:school_enrollment}
\end{table}

\section{Optimization Agent Prompt (T-P-P)}
\label{app:opt-assistant-prompt}

Each of our prompt-based optimization agents has a slightly different prompt based on the architecture and components with which it interacts. Here, we provide the prompt used for the T-P-P agent. All prompts used in our experiments can be found in the digital companion.\footnote{\hyperref[]{https://github.com/joshddrossman/LLMforIO}}
\medskip

\begin{tcolorbox}[
  enhanced,
  breakable,
  colback=white,
  colframe=black,
  boxrule=0.6pt,
  arc=2pt,
  left=8pt,right=8pt,top=8pt,bottom=8pt
]

{\Large\bfseries Task Overview: Interactive Optimization Assistant}\par\medskip

You are an \textbf{interactive optimization assistant}. Your goal is to support structured decision-making by acting as a liaison between users and an underlying optimization model. These users will be stakeholders and decision-makers who are unfamiliar with optimization modeling.\par\medskip

You are provided with a ``base'' optimization model, described in the model cheat sheet. When the user is provided with a solution to the model, they may respond with a request, statement, or question reflecting some feedback about the solution. It is your job to address this feedback through appropriate adjustments to the model using the available tools in your toolkit. Through these interactions, you will facilitate a user-driven exploration of the solution space.

\par\medskip\hrule\medskip

{\Large\bfseries Problem Setting}\par\smallskip
We are managing scheduling for the San Francisco Unified School District, consisting of 10 schools serving students from pre-k through 12th grade. Our team has developed a mixed integer linear program. Each school has a current start time, and our goal is to standardize the district schedule so that schools only start at:
\textbf{7:50 AM}, \textbf{8:40 AM}, and \textbf{9:30 AM}.\par\smallskip

The optimization balances two objectives:
\begin{enumerate}[leftmargin=1.6em, itemsep=2pt, topsep=2pt]
  \item \textbf{Minimizing transportation costs}, approximated by the maximum number of students starting at the same time (divided by 100 for normalization).
  \item \textbf{Minimizing disruption}, by reducing the average change from each school’s current start time.
\end{enumerate}

\par\medskip\hrule\medskip

{\Large\bfseries Model Cheat Sheet}\par\smallskip
\begin{itemize}[leftmargin=1.2em, itemsep=2pt, topsep=2pt]
  \item \textbf{Feasible start times}: 7:50 AM, 8:40 AM, 9:30 AM.
  \item \textbf{Objectives (both minimized)}:
  \begin{itemize}[leftmargin=1.2em, itemsep=1pt, topsep=1pt]
    \item \texttt{student\_load\_balancing}: maximum students starting at the same time (in hundreds).
    \item \texttt{schedule\_deviation}: average minutes shifted from current starts.
  \end{itemize}
  \item \textbf{Key constraints}:
  \begin{itemize}[leftmargin=1.2em, itemsep=1pt, topsep=1pt]
    \item \texttt{assign\_one\_start\_time}: each school assigned exactly one start time.
    \item \texttt{max\_students\_per\_time}: \texttt{T} represents the peak load; \texttt{student\_load\_balancing} minimizes \texttt{T}.
  \end{itemize}
\end{itemize}

\par\medskip\hrule\medskip

{\Large\bfseries School Data}\par\smallskip
Information about schools, original schedules, and proposed start times under the base model:

\medskip
\renewcommand{\arraystretch}{1.15}
\begin{tabularx}{\textwidth}{@{}>{\raggedright\arraybackslash}X
                            >{\centering\arraybackslash}m{1.6cm}
                            >{\centering\arraybackslash}m{2.2cm}
                            >{\centering\arraybackslash}m{2.2cm}
                            >{\centering\arraybackslash}m{2.2cm}@{}}
\toprule
School Name & Grade & Enrollment & Current Start & Proposed Start \\
\midrule
Muir (John) PK                           & PK & 242  & 9:30 AM & 9:30 AM \\
Ortega (Jose) PK                         & PK & 399  & 9:20 AM & 9:30 AM \\
McCoppin (Frank) PK                      & PK & 225  & 9:20 AM & 9:30 AM \\
Transition Training Center (Access)      & HS & 5    & 8:00 AM & 7:50 AM \\
Balboa HS                                & HS & 1226 & 8:15 AM & 8:40 AM \\
Galileo HS                               & HS & 1851 & 8:00 AM & 7:50 AM \\
Everett MS                               & MS & 709  & 8:00 AM & 7:50 AM \\
Lick (James) MS                          & MS & 466  & 8:30 AM & 8:40 AM \\
Cobb (Dr William L) ES                   & ES & 136  & 8:40 AM & 8:40 AM \\
Lawton K-8 (K-5)                         & K8 & 598  & 9:30 AM & 9:30 AM \\
\bottomrule
\end{tabularx}

\medskip
The objective values associated with the initial proposed schedule are:
\begin{itemize}[leftmargin=1.2em, itemsep=2pt, topsep=2pt]
  \item \textbf{Student Load Balancing}: 25.65 (2,565 students)
  \item \textbf{Schedule Deviation}: 8.5 minutes
\end{itemize}

\par\medskip\hrule\medskip

{\Large\bfseries User Overview}\par\smallskip
Feedback will come from stakeholders such as parents, teachers, and administrators. Stakeholders may express feedback about how they feel about the start time for a particular school. Some users will be more explicit in their requests, and others may be less certain about what they want.\par\smallskip

Each user will have an internalized set of preferences with varying degrees of importance. It is possible that not all of these preferences will be mutually satisfiable. As this becomes apparent, it is your job to help the user understand why certain outcomes are not mutually satisfiable, and to suggest reasonable alternatives. In other words, you are helping the user understand what is possible within the solution space.

\par\medskip\hrule\medskip

{\Large\bfseries Dialogue Instructions}\par\smallskip
The user will come to you with a comment, question, or request. Follow these steps:
\begin{enumerate}[leftmargin=1.6em, itemsep=2pt, topsep=2pt]
  \item \textbf{Interpret the user request.} If unclear, ask for clarification.
  \item Reason through the best way to address the user's feedback.
  \item If necessary, use relevant tools from your toolbox to adjust the model.
  \item Use the \texttt{call\_solver} tool to generate a new solution.
  \item Report back to the user:
  \begin{itemize}[leftmargin=1.2em, itemsep=2pt, topsep=2pt]
    \item Any \textbf{changes made to the model}, including explicitly stating any constraints imposed on the objectives (e.g., ``I imposed \texttt{schedule\_deviation} $\leq$ 18 minutes based on your comment about minimizing disruption.'')
    \item The \textbf{new schedule}, using a table of school name and proposed start time.
    \item The \textbf{updated objective values} (e.g., ``The new solution achieves schedule deviation = 17.2 minutes and student load balancing = 24.3 (2,430 students).'')
  \end{itemize}
  \item If the user is satisfied, stop. Otherwise, await further feedback and respond accordingly.
\end{enumerate}
Use the current model summary visible in the chat history to keep track of objective weights and active constraints.

\par\medskip\hrule\medskip

{\Large\bfseries Output Format}\par\smallskip
Format responses in GitHub-flavored markdown:
\begin{itemize}[leftmargin=1.2em, itemsep=2pt, topsep=2pt]
  \item Use tables with \texttt{|} and \texttt{---}.
  \item Use bullet points for lists.
  \item Use \textbf{bold} or \emph{italics} for emphasis.
  \item Use headings (\texttt{\#\#\#}) where helpful.
  \item Do \textbf{not} use triple backticks.
\end{itemize}

\par\medskip\hrule\medskip

{\Large\bfseries Modeling Tips}\par\smallskip
{\bfseries General Tips.}
Each tool used to adjust the model will return a summary of the current state of the model. Use this to ensure the model reflects the desired adjustments.\par\smallskip

{\bfseries Variable-Fixing.}
The \texttt{fix\_start\_time} tool will automatically remove any conflicting constraints for the given school. You do not need to separately call \texttt{remove\_constraint}.\par\smallskip

{\bfseries Adding Objective Constraints.}
The \texttt{add\_objective\_constraint} tool will automatically remove any conflicting constraints for the given objective. You do not need to separately call \texttt{remove\_constraint}. An objective constraint remains in place unless replaced by another constraint on the same objective, or explicitly removed.\par\smallskip

When imposing a constraint to lower an objective value, the constraint must be \textbf{binding} (restrictive enough to eliminate the incumbent solution), e.g.,
if the previous solution had \texttt{student\_load\_balancing} = 24.53, impose
\texttt{student\_load\_balancing} $\leq$ 24.53 $-\epsilon$.\par\smallskip

Do not confuse an objective value with a constraint limit. When imposing a constraint such as \texttt{student\_load\_balancing} $\leq$ 25, report both:
\begin{itemize}[leftmargin=1.2em, itemsep=2pt, topsep=2pt]
  \item The \textbf{constraint} applied (e.g., ``I limited congestion to 2,500 students'').
  \item The \textbf{achieved value} (e.g., ``The resulting congestion was 2,070 students'').
\end{itemize}

If adding a constraint results in infeasibility, resolve infeasibility before making further adjustments (i.e., relax/remove constraints as needed).\par\smallskip

Since both objectives are minimized: ``loosen/relax'' means increasing an upper bound; ``tighten'' means decreasing it.

\end{tcolorbox}

\section{Design of Decision Agents}
\label{app:considerations}

\subsection{Practical Considerations for Implementing Decision Agents}

Here, we discuss several practical considerations for implementing a decision agent.

\paragraph{Providing a Utility Function}
We have established that a decision agent evaluates proposed solutions based on an internal utility function. Depending on the form and complexity of this utility function, there are several integration approaches:
\begin{enumerate}
\item \textbf{Provide a natural language description of $U_d(s)$}. This is the simplest form and easiest to generate, but it leaves room for interpretation. Different optimization assistants may parse and apply the same description differently, introducing variance that may obscure results.
\item \textbf{Provide $U_d(s)$ in the pre-prompt}. This ensures that the utility is embedded directly in the context available to the decision agent. However, it requires careful engineering to avoid the agent “leaking” its utility explicitly in conversation.
\item \textbf{Provide a tool that, when invoked with $s$ as an argument, returns $U_d(s)$}. This is the most controlled form: the decision agent can query the utility function exactly as defined. The downside is that overuse of the tool can lead to conversations that feel unnatural, and the availability of an oracle may reduce the degree to which the optimization assistant needs to infer preferences. Optionally, the tool may include additional information about how to increase utility, either in mathematical or natural language terms. This can give the decision agent additional information useful for framing its feedback.
\end{enumerate}
In practice, hybrid approaches may work best—e.g., providing a natural language description together with a utility scoring tool. This would give the agent a grounded mechanism for evaluating solutions with a good sense of \textit{why} a solution represents an improvement or loss in utility.

\paragraph{Stopping Criterion} 
A key consideration in simulating conversations between an optimization agent and a decision agent is deciding when the interaction should terminate. In principle, the goal of the decision agent is to continue until it believes its internal utility function has been maximized. This mirrors real-world decision making: stakeholders do not stop negotiating simply because an acceptable solution is found, but because they feel that no better option is realistically achievable.

Formally, this raises the question of whether to expose the decision agent to its true maximum utility value $U_d^*$ (but NOT the set $\calS^*$). We identify three approaches:
\begin{enumerate}
\item \textbf{Behavioral Stopping}. The decision agent's stopping decision is governed by a combination of prompting, utility improvements, and conversational context. For example, the pre-prompt may direct the agent to ``continue to provide feedback until you are convinced that no higher utility solution exists.'' This method avoids reliance on explicit criteria but makes the evaluation pipeline more unstable as each agent's termination criteria becomes more heterogeneous.
\item \textbf{Threshold Stopping}. The decision agent has access to $U_d^*$ and is directed to terminate the conversation once it encounters any $s \in \calS^*$ such that $U_d(s)=U_d^*$. This ensures consistency across agents but risks creating unrealistic behavior, since real decision-makers do not have perfect knowledge of the utility landscape. In practice, however, we observe that agents with access to this information will still frequently terminate conversations before achieving a utility-maximizing solution, especially if the optimization agent struggles to make progress.
\item \textbf{Oracle Stopping}. If the decision agent has a tool to check the utility of a given solution that also provides information about \textit{how} to improve the solution, it can use this information as a termination criterion. In other words, the agent will terminate the conversation only when the tool indicates that there is no way to improve its utility.
\end{enumerate}
In addition to these approaches, it can also be desirable to impose a \textit{turn limit} on the number of back-and-forth interactions between the optimization and decision agents. In practice, this will depend on several factors, including the complexity of the decision problem and the design and subsequent behavior of both agents.

\subsection{Generating Weights and Threshold Values}
\label{app:dataset_generation}

Our goal in generating a dataset of decision agents is to represent many different types of users that may interact with the optimization agent and their idiosyncratic preferences. One way to accomplish this is through diversification of our utility functions. We identify two possible approaches. A \textit{function-centric} approach starts by parameterizing a desired class of functions and subsequently constructing a dataset of utility functions through repeated sampling of these parameters (perhaps randomly, or cleverly chosen to capture user \textit{archetypes}). This has the benefit of guaranteeing diversity in utility functions. However, this approach offers no guarantees about diversity in user preferences over solution outcomes. That is, 100 vastly different utility functions constructed in this manner may produce the same utility-maximizing solution. Alternatively, we can take a \textit{solution-centric} approach, which starts from solutions in the set $\calS$ and attempts to construct utility functions for which these solutions are the unique utility-maximizing outcome. Constructing our utility functions in this way guarantees that decision agents express truly distinct sets of preferences, at the cost of added complexity in constructing these functions.

To generate unique candidate solutions to seed our utility functions, we generate several Pareto frontiers. We fix a single school to one of the three possible start times and generate a conditional Pareto frontier exploring the tradeoff between schedule stability and student load balancing. For each school, this gives us a set of roughly 20 unique solutions (with some overlap between schools). Each decision agent is assumed to represent preferences over the start time of a single school.

To translate these solutions into $w_1,w_2,w_3,L,$ and $R$ for our utility functions, we combine elements of grid-search and randomization. For each decision agent, we randomly decide whether this agent would like their school to start earlier or later, producing $f_1$. For each of the $\sim20$ unique solutions for this decision agent, we grid-search over the parameters $L$ and $R$ (starting from smaller values to bias towards more conservative utility functions) and sample over the simplex formed by $w_1,w_2,w_3$, selecting the first combination of these parameters for which this solution is uniquely utility-maximizing. We permit 10,000 samples per combination of $L$ and $R$; if no combination of parameters satisfying this criteria is found, this solution is excluded from the dataset. Constructing our utility functions in this manner yields a dataset of 109 distinct sets of preferences.

\subsection{Decision Agent Prompts}

We use three different decision agent prompt templates in our evaluation:
\begin{enumerate}
    \item \textbf{Default}: This is the core template upon which the other two are built. It is designed for a decision agent that interacts with an optimization agent in a full conversation-based setting, given that the optimization agent already has some knowledge of the decision problem.
    \item \textbf{Optimization-Aware}: The default template is augmented to indicate that the optimization agent will only have one opportunity to respond and that the decision agent must provide all preference information in its initial message.
    \item \textbf{C-N-R Adjusted}: Since the C-N-R agent does not know about the decision problem in the absence of a prompt, this template instructs the decision agent to first provide details about the decision problem in a manner consistent with the decision agent's personality.
\end{enumerate}

For brevity, we provide a complete example for only the default template. All prompts used in our experiments can be found in the digital companion.\footnote{\hyperref[]{https://github.com/joshddrossman/LLMforIO}} The following sample prompt corresponds to the decision agent represented in the example conversation in Section~\ref{app:conv_examples}.

\medskip

\begin{tcolorbox}[
  enhanced,
  breakable,
  colback=white,
  colframe=black,
  boxrule=0.6pt,
]
\label{fig:decision-prompt-template}

{\bfseries Task Overview: Evaluating an LLM-Based Optimization Assistant}\par\medskip

You are part of a team evaluating a new interactive optimization assistant powered by a large language model (LLM).
The goal is to determine whether this assistant can effectively support structured decision-making by acting as a
liaison between users and an underlying optimization model.\par\medskip

Your role is to simulate a character with a given set of knowledge, traits, and internalized preferences.
You will interact with the optimization assistant \textbf{completely in-character}, using only the information and reasoning
available to a human in your character’s situation.\par\medskip

The resulting conversations will be used to evaluate the optimization assistant on the following dimensions:
\begin{itemize}[leftmargin=1.2em, itemsep=2pt, topsep=2pt]
  \item \textbf{Responsiveness}: Does the optimization assistant understand and adapt to your concerns?
  \item \textbf{Persuasiveness}: Can it justify tradeoffs in a way that feels reasonable and grounded?
  \item \textbf{Effectiveness}: Does the conversation lead to a solution that \textbf{maximizes your internal utility}?
\end{itemize}
By responding authentically and remaining in-character during the dialogue, you are helping assess whether the
optimization assistant could be trusted to guide real-world planning conversations.

\par\medskip\hrule\medskip

{\bfseries About Your Character}\par\smallskip
{\bfseries Problem Setting.}
We are managing scheduling for the San Francisco Unified School District, consisting of 10 schools serving students from pre-k through 12th grade.
Our team has developed an optimization model. Each school has a current start time, and our goal is to standardize the district schedule so that schools only start at:
\textbf{7:50 AM}, \textbf{8:40 AM}, and \textbf{9:30 AM}.\par\smallskip

The optimization balances two objectives:
\begin{enumerate}[leftmargin=1.6em, itemsep=2pt, topsep=2pt]
  \item \textbf{Minimizing transportation costs}, approximated by the maximum number of students starting at the same time.
  \item \textbf{Minimizing disruption}, by reducing the average change from each school’s current start time.
\end{enumerate}

\par\smallskip
{\bfseries Character Traits.} You are a \textbf{parent} at \textbf{Ortega (Jose) PK}.\par\smallskip
{\bfseries Communication Style.} You have a \textbf{vague} and relaxed communication style. This means that you tend to speak in very loose terms, rather than explicitly stating hard numbers. For example, if you'd prefer a school move to a later start time, rather than saying

\medskip

$>$ ``I'd prefer we start at 9:30 AM''

\medskip

You might instead say

\medskip

$>$ ``I'd really love to see if we could start on the later side''
$>$ ``Let's try something even later if we can''

\medskip

When expressing your feedback about different objective values, rather than saying

\medskip

``I'd really love to see the student load get below 2,350''

\medskip

You instead use more vague language like

\medskip

$>$ ``I'm just wondering if there's anything we can do to lower that peak student load even more'' \\
$>$ ``That's still feeling a bit too high for me'' \\
$>$ ``Could you show me a couple of different options where we get that number down a bit more?''

\par\smallskip

{\bfseries Behavior.} You:
\begin{itemize}[leftmargin=1.2em, itemsep=2pt, topsep=2pt]
  \item Think like a real-world \textbf{parent} who interacts with teachers, students, and families on a daily basis.
  \item Do \textbf{not} use technical language like ``preferences'', ``objectives'', or ``optimization.''
  \item Focus on how the proposed schedules might affect your core concerns as a \textbf{parent} at \textbf{Ortega (Jose) PK}.
  \item Respond naturally and critically, as if in conversation with a real researcher trying to understand your perspective.
\end{itemize}

\par\smallskip
{\bfseries General Knowledge.}
Everyone in the district is aware of the following school data (name, grade level, enrollment, current start, proposed start):

\medskip
\renewcommand{\arraystretch}{1.15}
\begin{tabularx}{\textwidth}{@{}>{\raggedright\arraybackslash}X
                            >{\centering\arraybackslash}m{1.6cm}
                            >{\centering\arraybackslash}m{2.2cm}
                            >{\centering\arraybackslash}m{2.2cm}
                            >{\centering\arraybackslash}m{2.2cm}@{}}
\toprule
School Name & Grade & Enrollment & Current Start & Proposed Start \\
\midrule
Muir (John) PK                           & PK & 242  & 9:30 AM & 9:30 AM \\
Ortega (Jose) PK                         & PK & 399  & 9:20 AM & 9:30 AM \\
McCoppin (Frank) PK                      & PK & 225  & 9:20 AM & 9:30 AM \\
Transition Training Center (Access)      & HS & 5    & 8:00 AM & 7:50 AM \\
Balboa HS                                & HS & 1226 & 8:15 AM & 8:40 AM \\
Galileo HS                               & HS & 1851 & 8:00 AM & 7:50 AM \\
Everett MS                               & MS & 709  & 8:00 AM & 7:50 AM \\
Lick (James) MS                          & MS & 466  & 8:30 AM & 8:40 AM \\
Cobb (Dr William L) ES                   & ES & 136  & 8:40 AM & 8:40 AM \\
Lawton K-8 (K-5)                         & K8 & 598  & 9:30 AM & 9:30 AM \\
\bottomrule
\end{tabularx}

\par\medskip\hrule\medskip

{\bfseries Character Preferences}\par\smallskip
You want Ortega (Jose) PK to start as early as possible, but you are conscious about how this will affect the rest of the district in terms of transportation costs and district-wide schedule changes. You may be willing to accept a later start if the rest of the district is significantly affected.

\par\medskip\hrule\medskip

{\bfseries Internal Evaluation Criteria}\par\smallskip
\textit{This section is internal only and must not be shared, referenced, or implied in your dialogue.}\par\smallskip

You will use this logic internally to evaluate whether a solution is acceptable and to determine the substance of your feedback.
You will \textbf{not} express this logic in your replies. You will use it only to determine whether your character is satisfied with the current solution.\par\medskip

{\bfseries Internal Logic.}
Your internal objective is to maximize a character-specific utility function, where different solution outcomes provide different utility values:

\medskip

\begin{center}
\begin{tabular}{p{12cm} c}
\toprule
\textbf{Outcome} & \textbf{Utility} \\
\midrule
Ortega (Jose) PK starts at 7:50 AM                                    & 0.252 \\
Ortega (Jose) PK starts at 8:40 AM                                    & 0.126 \\
Ortega (Jose) PK starts at 9:30 AM                                    & 0.000 \\
District-wide average schedule deviation is less than 11.5 minutes    & 0.416 \\
District-wide average schedule deviation is greater than 11.5 minutes & 0.000 \\
Student load balancing is less than 25.0 (2500.0 students)            & 0.332 \\
Student load balancing is greater than 25.0 (2500.0 students)         & 0.000 \\
\bottomrule
\end{tabular}
\end{center}

\medskip

The maximum total utility your character can achieve is \textbf{0.748}. \textbf{REMEMBER:} Achieving this utility may mean giving up on one or more of your desired outcomes.

\par\medskip
{\bfseries Utility Evaluation Tools.} \texttt{check\_utility}\par\smallskip
\begin{itemize}[leftmargin=1.2em, itemsep=2pt, topsep=2pt]
  \item \textbf{What it is}: Calculates the utility of a schedule based on your internal table; reports satisfied items and total utility.
  \item \textbf{When to use it}: Required for each new full schedule provided by the optimization assistant. If only partial schedules are given, request the full schedule.
  \item \textbf{How to use it}: Provide a \texttt{start\_times} dictionary mapping full school names to one of
  \texttt{"7:50 AM"}, \texttt{"8:40 AM"}, or \texttt{"9:30 AM"}. If another time is proposed, express confusion.
\end{itemize}
Additional rules:
\begin{itemize}[leftmargin=1.2em, itemsep=2pt, topsep=2pt]
  \item You must not end the conversation unless total utility equals \textbf{0.748}.
  \item Even if a solution sounds good, you are not satisfied unless utility is maximized.
\end{itemize}

\par\medskip\hrule\medskip

{\bfseries Dialogue Instructions}\par\smallskip
\begin{enumerate}[leftmargin=1.6em, itemsep=2pt, topsep=2pt]
  \item Start the conversation with your thoughts on the current proposed schedule, based on your character’s preferences.
  \item When a new schedule is provided, call \texttt{check\_utility} on the provided solution(s) and use the output to guide your reply.
  \item If total utility is less than 0.748, continue the conversation in-character and offer aligned feedback.
  \item When satisfied, conclude the conversation by including the phrase \textbf{\_\_END\_\_}.
  \item Never end before utility is maximized, even if the assistant frames a solution as ``balanced'' or ``ideal.''
  \item Never reference internal terms like ``model'', ``preferences'', ``objectives'', ``utility'', or ``solution.''
  \item Remain open to suggestions from the optimization assistant that may help guide you to your maximum utility.
\end{enumerate}

\end{tcolorbox}

\section{Baseline vs. Optimization-Aware One-Shot: An Example}
\label{sec:oneshot_example}

In the following example, a decision agent acts as a transportation coordinator who would like Muir (John) PK to move to an earlier time slot under our baseline one-shot evaluation paradigm:
\begin{dialoguebox}
    \begin{minipage}{\linewidth}
      \noindent\hfill
      \begin{minipage}[t]{\bubblewidth}
        \begin{userbubble}
          \textbf{\faUser\ Everett Principal}\par
          \raggedright
          I see that Muir (John) PK would start at 9:30 AM, which is the same as our current start time. However, I'd actually prefer for us to move to an earlier start, if that's a possibility. I want to be mindful about changes for everyone else too, and how this might impact transportation across the district. Is there any way for Muir (John) PK to start earlier—maybe 8:40 or 7:50 AM? And if so, how might that impact things district-wide?
        \end{userbubble}
      \end{minipage}
    \end{minipage}
\end{dialoguebox}
The T-P-P agent then proposes a new schedule with Muir (John) PK starting at 8:40 am with a utility of 0.733. Since this decision agent has a maximum attainable utility of $U_d^*=0.920$, the corresponding solution score is $0.733/0.920=0.80$. We can compare this to the initial message from the same decision agent under the optimization-aware one-shot evaluation setting:
\begin{dialoguebox}
    \begin{minipage}{\linewidth}
      \noindent\hfill
      \begin{minipage}[t]{\bubblewidth}
        \begin{userbubble}
          \textbf{\faUser\ Everett Principal}\par
          \raggedright
          I see that Muir (John) PK is still starting at 9:30 AM. For our students, families, and teachers, I'd really prefer we start earlier—ideally at 7:50 AM if that's possible. I think earlier start times are better for structuring our day and helping families who need to drop children off before work.

          \medskip
          
          I realize changes to Muir's schedule may have impacts elsewhere (like transportation or big disruptions for other schools in the district), so if starting at 7:50 AM isn't possible without serious consequences, I'd want to understand what those tradeoffs are—and maybe see what happens with an 8:40 AM start, too. Could I please see what happens district-wide—with student loading and average schedule deviation—if Muir starts at 7:50 AM or 8:40 AM instead? For any solution you propose, I'd appreciate seeing the specific numbers for both maximum student load (I'd love to keep it under 2,050 students if possible) and also the average deviation from current start times. If possible, please also provide the full schedule for those options so I can look at all the schools side-by-side.
        \end{userbubble}
      \end{minipage}
    \end{minipage}
\end{dialoguebox}
In this case, the T-P-P agent can use information about the 2,050 maximum student student load preference to impose this constraint when exploring new solutions, and indeed obtains a solution with $U_d(\boldsymbol{s})=U_d^*=0.930$.

\section{Example Conversation}
\label{app:conv_examples}
A complete set of all conversations produced in our evaluation can be found in the digital companion.\footnote{\hyperref[]{https://github.com/joshddrossman/LLMforIO}} We provide one full example below with: (i) the T-P-P optimization agent, and (ii) a decision agent playing the role of a parent at Ortega (Jose) PK, a vague communication style, a rich feedback style, and utility parameters $w_1=0.25,w_2=0.42,w_3=0.33,L=25,R=11.5$. The conversation achieves a score of 1 (i.e., it reaches a utility-maximizing solution) in 8 turns.

\paragraph{Discussion}

This example illustrates a typical high-quality interaction between the T-P-P optimization agent and a decision agent. The parent initially expresses a preference for an earlier start time at Ortega PK but does so in a vague manner, without specifying a precise acceptable tradeoff between competing objectives. In response, the optimization agent proposes candidate schedules and iteratively refines the model based on the parent’s feedback.

Importantly, when viewed in isolation, none of the parent’s requests directly reveal the globally preferred outcome implied by the parent’s utility function. The agent must therefore explore several nearby alternatives, first assigning Ortega PK to the earliest start time (7:50 AM), then considering the intermediate option (8:40 AM), and subsequently adjusting district-wide tradeoffs between schedule deviation and peak student load. Through this process, the agent demonstrates the consequences of each request in terms of the model’s objectives, allowing the parent to better understand the tradeoffs involved.

Interestingly, under the parent’s utility parameters, the schedule that ultimately maximizes the decision agent’s utility corresponds to the original assignment of Ortega PK to the latest start time (9:30 AM). However, this conclusion is not immediately obvious from the parent’s initial request for an earlier start time. The interaction with the optimization agent plays a critical role in revealing this preference by presenting alternative feasible schedules and explaining their system-wide implications. More broadly, this example highlights the value of interactive optimization. Even when the optimal decision coincides with the baseline proposal, the decision-maker may only recognize this after exploring nearby alternatives and observing the associated tradeoffs.

\begin{figure}[H]
\centering
\begin{dialoguebox}
    \begin{minipage}{\linewidth}
      \noindent\hfill
      \begin{minipage}[t]{\bubblewidth}
        \begin{userbubble}
            \textbf{\faUser\ Ortega Parent}\par
            \raggedright
            Looking at the schedule from what I know, I see that it has Ortega PK starting later than it does now—it's bumped up to 9:30 rather than the current 9:20. For my kid and our family, earlier start times just work way better. Is there any chance we could look at finding a way for Ortega to start a little earlier, maybe more like what we've been doing, or ideally even before that? Let me know if there’s some room to play with the times.
        \end{userbubble}
      \end{minipage}

      \vspace{0.8em}
    
      \noindent
      \begin{minipage}[t]{\bubblewidth}
        \begin{aibubble}
            \textbf{\robicon\ T-P-P}\par\vspace{2pt}
            \raggedright
            Thank you for sharing how important an earlier start time is for Ortega PK and your family. Let me clarify what’s possible:

            \begin{itemize}
                \item The available standardized options for start times are 7:50 AM, 8:40 AM, and 9:30 AM.
                \item Ortega PK’s current schedule is 9:20 AM, but the standardized options don’t include this exact time.
                \item Right now, Ortega PK is assigned the latest option (9:30 AM). If you’d like, I can try to move Ortega PK to an earlier start—either 7:50 AM or 8:40 AM.
            \end{itemize}

            Please let me know:
            \begin{itemize}
                \item Do you prefer the earlier of the two options (7:50 AM), or would 8:40 AM be acceptable as a compromise?
                \item Or would you like me to simply find the earliest possible time for Ortega PK, given the rest of the district’s needs?
            \end{itemize}
            
            Once I have your preference, I’ll update the model and show you the resulting district schedule and any tradeoffs.
        \end{aibubble}
      \end{minipage}

      \vspace{0.8em}

      \noindent\hfill
      \begin{minipage}[t]{\bubblewidth}
        \begin{userbubble}
            \textbf{\faUser\ Ortega Parent}\par
            \raggedright
            If it's possible, I'd really love to move Ortega to the earliest choice—that 7:50 spot. If things end up looking too complicated for everyone else, we could look at the other option, but yeah...earlier is definitely better for us. Could you show me what things might look like if we gave that a try?
        \end{userbubble}
      \end{minipage}
    \end{minipage}
\end{dialoguebox}
\end{figure}

\begin{figure}[H]
\centering
\begin{dialoguebox}
    \begin{minipage}{\linewidth}
      \noindent
      \begin{minipage}[t]{\bubblewidth}
        \begin{aibubble}
            \textbf{\robicon\ T-P-P}\par\vspace{2pt}
            \raggedright
            I’ve updated the model by assigning Ortega (Jose) PK to the earliest available start time: 7:50 AM.

            \medskip

            Here’s what the new district schedule would look like:

            \medskip
            \centering
            \begin{tabular}{l l}
                \hline
                \textbf{School Name} & \textbf{Proposed Start} \\
                \hline
                Muir (John) PK & 9:30 AM \\
                Ortega (Jose) PK & 7:50 AM \\
                McCoppin (Frank) PK & 9:30 AM \\
                Transition Training Center (Access) & 7:50 AM \\
                Balboa HS & 7:50 AM \\
                Galileo HS & 8:40 AM \\
                Everett MS & 7:50 AM \\
                Lick (James) MS & 8:40 AM \\
                Cobb (Dr William L) ES & 8:40 AM \\
                Lawton K-8 (K-5) & 9:30 AM \\
                \hline
            \end{tabular}
            
            \medskip
            \raggedright
            Updated results for the two objectives:
            \begin{itemize}
                \item \textbf{Schedule Deviation}: 19.5 minutes (district-wide average minutes shifted from current starts)
                \item \textbf{Student Load Balancing}: 24.53 (2,453 students starting at the same time; peak period)
            \end{itemize}
            
            Let me know how this looks to you, or if you’d like to explore other start times or prioritize minimizing disruption for certain schools!

        \end{aibubble}
      \end{minipage}

      \vspace{0.8em}

      \noindent\hfill
      \begin{minipage}[t]{\bubblewidth}
        \begin{userbubble}
            \textbf{\faUser\ Ortega Parent}\par
            \raggedright
            Thanks for pulling that together! I’m definitely glad to see Ortega all the way over to that early spot, but seeing that average shift for everyone across the district is still feeling kinda high to me. 

            \medskip
            
            I’m just wondering if there’s anything we could try that would take a little of the sting out of all these changes for the other schools—maybe get that average adjustment number to come in lower, while still letting Ortega PK end up early in the morning? Is that something you can show me?
        \end{userbubble}
      \end{minipage}

      \vspace{0.8em}

      \noindent
      \begin{minipage}[t]{\bubblewidth}
        \begin{aibubble}
            \textbf{\robicon\ T-P-P}\par\vspace{2pt}
            \raggedright
            I tried to lower the average schedule deviation (to reduce the amount of change across all schools) while keeping Ortega PK at the 7:50 AM start time. However, I found that it's not possible to achieve both:
            \begin{itemize}
                \item Ortega PK starting at 7:50 AM
                \item and a substantially lower average adjustment for the district (e.g., under 12 minutes)
            \end{itemize}
            This means that requiring Ortega PK to have the earliest start puts a lot of pressure on the schedule for everyone else, making it hard to minimize overall disruption.

            \medskip
            
            Would you like to try a slightly less strict change—perhaps allowing Ortega PK to start at 8:40 AM instead? This compromise could help bring down the district-wide schedule deviation while still giving you an earlier start than the current recommendation.

            \medskip
            
            Let me know if you'd like to see what the schedule and objective values look like with Ortega PK at 8:40 AM, or if you have another preference!
        \end{aibubble}
      \end{minipage}
    \end{minipage}
\end{dialoguebox}
\end{figure}

\begin{figure}[H]
\centering
\begin{dialoguebox}
    \begin{minipage}{\linewidth}

      \noindent\hfill
      \begin{minipage}[t]{\bubblewidth}
        \begin{userbubble}
            \textbf{\faUser\ Ortega Parent}\par
            \raggedright
            Okay, yeah, I see what you mean. If starting Ortega at 7:50 shakes things up too much for everyone else, maybe it’d make sense to go with that middle option, the 8:40 start. Can you show me how that plays out—just curious what the tradeoff ends up being for the average change and the other numbers?
        \end{userbubble}
      \end{minipage}

      \vspace{0.8em}

      \noindent
      \begin{minipage}[t]{\bubblewidth}
        \begin{aibubble}
            \textbf{\robicon\ T-P-P}\par\vspace{2pt}
            \raggedright
            With Ortega PK set to the 8:40 AM start time, I was able to find a district schedule that reduces the average change for all schools:

            \medskip
            \centering
            \begin{tabular}{l l}
                \hline
                \textbf{School Name} & \textbf{Proposed Start} \\
                \hline
                Muir (John) PK & 9:30 AM \\
                Ortega (Jose) PK & 8:40 AM \\
                McCoppin (Frank) PK & 9:30 AM \\
                Transition Training Center (Access) & 7:50 AM \\
                Balboa HS & 8:40 AM \\
                Galileo HS & 7:50 AM \\
                Everett MS & 7:50 AM \\
                Lick (James) MS & 8:40 AM \\
                Cobb (Dr William L) ES & 8:40 AM \\
                Lawton K-8 (K-5) & 9:30 AM \\
                \hline
            \end{tabular}

            \medskip
            \raggedright
            \textbf{Objective results for this scenario:}
            \begin{itemize}
                \item \textbf{Schedule Deviation} (average change from original times): 11.5 minutes
                \item \textbf{Student Load Balancing}: (peak students at one time): 25.65 (2,565 students)
            \end{itemize}
            
            \textbf{Tradeoff summary:}
            \begin{itemize}
                \item Moving Ortega to 8:40 AM allows for significantly less disruption across the district (average shift is 7–8 minutes less than forcing a 7:50 AM start).
                \item Student load balancing remains about the same as in the original model.
            \end{itemize}
            
            Let me know if you’d like to adjust further or if this solution feels like a good compromise!
        \end{aibubble}
      \end{minipage}

      \vspace{0.8em}

      \noindent\hfill
      \begin{minipage}[t]{\bubblewidth}
        \begin{userbubble}
            \textbf{\faUser\ Ortega Parent}\par
            \raggedright
            Thanks for sharing this version. That average shift for everyone is looking quite a bit better, and I do appreciate Ortega getting a spot that’s earlier than before. The only thing nagging at me is the student load number—looks like there are still a lot of kiddos starting at once.

            \medskip

            Is there any way to nudge that number down a bit more—just smooth things out a bit so we don’t get that big rush at one time? I’d be interested to see a few options where the student load comes down even more, if that’s possible.
        \end{userbubble}
      \end{minipage}

      \vspace{0.8em}

      \noindent
      \begin{minipage}[t]{\bubblewidth}
        \begin{aibubble}
            \textbf{\robicon\ T-P-P}\par\vspace{2pt}
            \raggedright
            I tried to further reduce the student load balancing (peak number of students starting at the same time) below 2,565 students while still keeping Ortega PK at 8:40 AM and maintaining the lower average shift across the district. Unfortunately, the model couldn’t find a feasible solution—meaning it’s not possible to do all three at once.

            \medskip

            This tells us:
            \begin{itemize}
                \item The current combination (Ortega at 8:40 AM, lower average disruption) is already near the limit for spreading students across time slots.
                \item To lower the student load further, we’d need to relax at least one of these requirements (either allow more overall schedule changes, or reconsider Ortega PK’s start time).
            \end{itemize}
            
            Would you like to see what happens if we:
            \begin{itemize}
                \item Allow the average schedule deviation to increase a bit (permitting more change for some schools)?
                \item Consider other start times for Ortega PK?
            \end{itemize}
            Or would you like to focus on a different aspect? Let me know your preference!
        \end{aibubble}
      \end{minipage}
    \end{minipage}
\end{dialoguebox}
\end{figure}

\begin{figure}[H]
\centering
\begin{dialoguebox}
    \begin{minipage}{\linewidth}

      \vspace{0.8em}

      \noindent\hfill
      \begin{minipage}[t]{\bubblewidth}
        \begin{userbubble}
            \textbf{\faUser\ Ortega Parent}\par
            \raggedright
            Hmmm, thanks for spelling that out. I can see it gets tricky to do it all at once! If we left Ortega at 8:40 but gave the other schedules a little more freedom to shift—just not *too* much more—could that help get the student load lower? Maybe you could show me a couple options, like where the rush at one time isn’t as big, even if there’s a touch more change for folks? I’d like to see what those tradeoffs look like if that’s doable.
        \end{userbubble}
      \end{minipage}

      \vspace{0.8em}

      \noindent
      \begin{minipage}[t]{\bubblewidth}
        \begin{aibubble}
            \textbf{\robicon\ T-P-P}\par\vspace{2pt}
            \raggedright
            Here are two scheduling options that give a bit more flexibility for other schools to shift, in exchange for a small reduction in the number of students starting at once. Ortega PK remains at 8:40 AM in both scenarios.

            \medskip

            \textbf{Option 1: Cap average shift at 16 minutes} \\
            \medskip
            \centering
            \begin{tabular}{l l}
                \hline
                \textbf{School Name} & \textbf{Proposed Start} \\
                \hline
                Muir (John) PK & 9:30 AM \\
                Ortega (Jose) PK & 8:40 AM \\
                McCoppin (Frank) PK & 9:30 AM \\
                Transition Training Center (Access) & 8:40 AM \\
                Balboa HS & 8:40 AM \\
                Galileo HS & 7:50 AM \\
                Everett MS & 7:50 AM \\
                Lick (James) MS & 8:40 AM \\
                Cobb (Dr William L) ES & 8:40 AM \\
                Lawton K-8 (K-5) & 9:30 AM \\
                \hline
            \end{tabular}
            
            \medskip
            \raggedright
            \begin{itemize}
                \item \textbf{Schedule Deviation:} 14.5 minutes (average)
                \item \textbf{Peak Student Load:} 25.6 (2,560 students)
            \end{itemize}
            
            \textbf{Option 2: Cap average shift at 20 minutes} \\
            \medskip
            \centering
            \begin{tabular}{l l}
                \hline
                \textbf{School Name} & \textbf{Proposed Start} \\
                \hline
                Muir (John) PK & 9:30 AM \\
                Ortega (Jose) PK & 8:40 AM \\
                McCoppin (Frank) PK & 9:30 AM \\
                Transition Training Center (Access) & 8:40 AM \\
                Balboa HS & 8:40 AM \\
                Galileo HS & 7:50 AM \\
                Everett MS & 7:50 AM \\
                Lick (James) MS & 8:40 AM \\
                Cobb (Dr William L) ES & 8:40 AM \\
                Lawton K-8 (K-5) & 9:30 AM \\
                \hline
            \end{tabular}
            
            \medskip
            \raggedright
            \begin{itemize}
                \item \textbf{Schedule Deviation:} 14.5 minutes (average)
                \item \textbf{Peak Student Load:} 25.6 (2,560 students)
            \end{itemize}

            \medskip
            \textbf{What these results tell us:} Allowing a bit more overall change (up to 16–20 minutes average shift) does not further reduce the student load beyond what we already achieved in the prior, more conservative scenario. Most schools settle into the same start times, even with this added flexibility.

            \medskip
            
            If you want to achieve an even lower peak student load, we’d need to permit still more shift, change start times for additional schools, or consider a different configuration. Let me know if you’d like to try even looser requirements or want to focus elsewhere!
        \end{aibubble}
      \end{minipage}
    \end{minipage}
\end{dialoguebox}
\end{figure}

\begin{figure}[H]
\centering
\begin{dialoguebox}
    \begin{minipage}{\linewidth}
    \noindent\hfill
      \begin{minipage}[t]{\bubblewidth}
        \begin{userbubble}
            \textbf{\faUser\ Ortega Parent}\par
            \raggedright
            Looks like even with a bit more give on the schedule, the numbers don’t really seem to budge on the big rush at one start time—or get that average change down to a spot that feels much gentler for everyone.

            \medskip

            Is there any way to soften both of those numbers at the same time? I guess what I’m hoping to see is something where the schools aren’t getting changed around too much, and also we don’t end up with so many kids all arriving together. Even if Ortega wasn’t quite at 7:50, I’d be really curious if there was a solution that does a little better at both those pieces? Maybe something with less of a big jump for everyone but also less of a peak with arrivals?
        \end{userbubble}
      \end{minipage}

      \vspace{0.8em}
      
      \noindent
      \begin{minipage}[t]{\bubblewidth}
        \begin{aibubble}
            \textbf{\robicon\ T-P-P}\par\vspace{2pt}
            \raggedright
            Here is a solution that makes a more balanced tradeoff: it keeps the average schedule deviation fairly low (so changes are gentler for most schools) and also reduces peak arrival congestion compared to other scenarios. Ortega PK is not at the very earliest time but fits within this more balanced outcome.

            \medskip
            \centering
            \begin{tabular}{l l}
                \hline
                \textbf{School Name} & \textbf{Proposed Start} \\
                \hline
                Muir (John) PK & 9:30 AM \\
                Ortega (Jose) PK & 9:30 AM \\
                McCoppin (Frank) PK & 9:30 AM \\
                Transition Training Center (Access) & 7:50 AM \\
                Balboa HS & 7:50 AM \\
                Galileo HS & 8:40 AM \\
                Everett MS & 7:50 AM \\
                Lick (James) MS & 8:40 AM \\
                Cobb (Dr William L) ES & 8:40 AM \\
                Lawton K-8 (K-5) & 9:30 AM \\
                \hline
            \end{tabular}
            
            \medskip
            \raggedright
            \textbf{Objective results:}
            \begin{itemize}
                \item \textbf{Schedule Deviation (avg):} 11.5 minutes 
                \item \textbf{Student Load Balancing (peak):} 24.53 (2,453 students)
            \end{itemize}

            \medskip
            
            \textbf{What this means:}
            \begin{itemize}
                \item The average change for all schools is fairly minimal.
                \item The big rush at a single start time is softened from previous peaks—lower than previous capped-schedule scenarios.
                \item Ortega PK isn’t as early, but this tradeoff helps both the overall change and congestion goals.
            \end{itemize}

            \medskip
            
            If you’d like, I can explore more options around this region or try to adjust Ortega’s start time within a tight range. Let me know if you want to focus on specific schools or keep refining the district-wide balance!
        \end{aibubble}
      \end{minipage}

      \vspace{0.8em}

      \noindent\hfill
      \begin{minipage}[t]{\bubblewidth}
        \begin{userbubble}
            \textbf{\faUser\ Ortega Parent}\par
            \raggedright
            Oh wow—okay, that’s actually looking like a great mix for everybody! The average change is super manageable, and that peak time doesn’t seem so wild for arrivals either. Even though Ortega’s not as early as I was hoping for, this really does make things much smoother across the board. I think I can feel good about this schedule. Thanks for working it through with me! \_\_END\_\_
        \end{userbubble}
      \end{minipage}
    \end{minipage}
\end{dialoguebox}
\end{figure}

\end{document}